\documentclass[sigconf]{acmart} 
\usepackage[utf8]{inputenc} % allow utf-8 input
\usepackage[T1]{fontenc}
\usepackage{varwidth} 
\usepackage{hyperref}       % hyperlinks
\usepackage{url}            % simple URL typesetting
\usepackage{booktabs}       % professional-quality tables
\usepackage{amsfonts}       % blackboard math symbols
\usepackage{nicefrac}       % compact symbols for 1/2, etc.
\usepackage{microtype}      % microtypography
\usepackage{xcolor}         % colors
\usepackage{soul}

\usepackage{amsmath} % for mathematical formulas
\usepackage{graphicx} % for including figures
\usepackage{subcaption}
\usepackage{natbib}
\usepackage{enumitem}
\usepackage{multirow}
\usepackage{bbm}
\usepackage{nth}
\usepackage{algorithm} 
\usepackage{algpseudocode} 
\usepackage{bm}
\usepackage{etoolbox}
\usepackage{makecell} 

\usepackage{listings}
\usepackage{comment}

\copyrightyear{2026}
\acmYear{2026}
\setcopyright{cc}
\setcctype{by}
\acmConference[IUI '26]{31st International Conference on Intelligent User Interfaces}{March 23--26, 2026}{Paphos, Cyprus}
\acmBooktitle{31st International Conference on Intelligent User Interfaces (IUI '26), March 23--26, 2026, Paphos, Cyprus}
\acmDOI{10.1145/3742413.3789078}
\acmISBN{979-8-4007-1984-4/2026/03}

\newcount\Comments  % 0 suppresses notes to selves in text
\newcommand{\kibitz}[2]{\ifnum\Comments=1{\color{#1}{#2}}\fi}
\definecolor{english}{rgb}{0.0, 0.5, 0.0}

\DeclareMathOperator*{\argmax}{\arg\!\max}

\begin{document}

\title{Strategic Tradeoffs Between Humans and AI in Multi-Agent Bargaining}

% The \author macro works with any number of authors. There are two commands
% used to separate the names and addresses of multiple authors: \And and \AND.
%
% Using \And between authors leaves it to LaTeX to determine where to break the
% lines. Using \AND forces a line break at that point. So, if LaTeX puts 3 of 4
% authors names on the first line, and the last on the second line, try using
% \AND instead of \And before the third author name.

\author{Crystal Qian}
\authornote{Both authors contributed equally.}
\affiliation{%
\institution{Google DeepMind}
\city{Mountain View}
\state{CA}
\country{USA}
}
\email{cjqian@google.com}

\author{Kehang Zhu}
\authornotemark[1]
\affiliation{%
\institution{Harvard University}
\city{Cambridge}
\state{MA}
\country{USA}
}
\email{kehang\_zhu@g.harvard.edu}

\author{John J. Horton}
\affiliation{%
\institution{MIT \& NBER}
\city{Cambridge}
\state{MA}
\country{USA}
}
\email{jjhorton@mit.edu}

\author{Benjamin S. Manning}
\affiliation{%
\institution{MIT}
\city{Cambridge}
\state{MA}
\country{USA}
}
\email{bmanning@mit.edu}

\author{Vivian Tsai}
\affiliation{%
\institution{Google DeepMind}
\city{Mountain View}
\state{CA}
\country{USA}
}
\email{vivtsai@google.com}

\author{James Wexler}
\affiliation{%
\institution{Google DeepMind}
\city{Mountain View}
\state{CA}
\country{USA}
}
\email{jwexler@google.com}

\author{Nithum Thain}
\affiliation{%
\institution{Google DeepMind}
\city{Mountain View}
\state{CA}
\country{USA}
}
\email{nthain@google.com}

\begin{abstract}
Markets increasingly accommodate large language models (LLMs) as autonomous decision-making agents.
As this transition occurs, it becomes critical to evaluate how these agents behave relative to their human and task-specific statistical predecessors. In this work, we present results from an empirical study comparing humans (N=216), multiple frontier LLMs, and customized Bayesian agents in dynamic multi-player bargaining games under identical conditions. 
Bayesian agents extract the highest surplus with aggressive trade proposals that are frequently rejected. 
Humans and LLMs achieve comparable aggregate surplus within their groups, but exhibit different trading strategies.
LLMs favor conservative, concessionary proposals that are usually accepted by other LLMs, while humans propose trades that are consistent with fairness norms but are more likely to be rejected.
These findings highlight that performance parity---a common benchmark in agent evaluation---can mask substantive procedural differences in \emph{how} LLMs behave in complex multi-agent interactions.

\end{abstract}


% CCS Concepts (choose weights as you like; 500=primary, 300=secondary)

\maketitle

\section{Introduction}

For actions with real social or economic consequences, humans have traditionally been the decision-makers, while statistical models have handled narrowly-scoped, well-defined tasks. As large language models (LLMs) have advanced, they have begun to upend this paradigm  \citep{GPTsAreGPTs2024,Shahidi2025Coasean}, exhibiting strong zero- and few-shot performance across many domains~\citep{brown2020language}. 
For example, call-center inquiries were once handled solely by human agents, then fronted by scripts and decision trees, and are now increasingly triaged by LLMs before reaching humans \citep{brynjolfsson2023generative}. Similar shifts are emerging across other interpersonal tasks. LLMs are used to negotiate contracts \citep{van2022walmart}, coordinate trade \citep{Corvin2024alibaba}, and facilitate e-commerce \citep{PYMNTS2025openai, gaarlandt2022shop}. This proliferation continues even though LLMs can fail to generalize in unexpected ways \citep{vafa2024general}. As they take on higher-stakes decisions, understanding their behavior relative to humans and previous generations of statistical models becomes essential---not only in terms of outcomes, but also in terms of alignment with procedural strategies and norms. 

Evaluating these tradeoffs requires assessing different types of agents in comparable yet sufficiently complex settings. This is difficult in practice. Across real-world applications, populations, incentives, and interfaces vary, making it hard to attribute observed differences to the agent rather than the setting. 
Static QA-style benchmarks---still the dominant LLM evaluation paradigm \citep{hendrycks2021mmlu}---fail to capture the dynamic, multi-agent settings that characterize many real-world interactions.
Indeed, recent work has called for more realistic evaluation environments that reflect the social and strategic complexities of the real-world \citep{raman2024steer, Goktas2025stratFoundation}.

\newpage

We address this challenge by introducing an incentivized, multi-player bargaining game. The game is designed to enable controlled comparisons among LLMs, humans, and bespoke statistical agents under identical conditions. In our setting, players are endowed with colored chips associated with fixed, private valuations, and they negotiate trades to maximize individual surplus, defined as the value of their final chip bundle minus the value of their initial endowment. Crucially, preferences are fully induced \citep{smith1976induced}: endowments and valuations are randomly drawn from a known distribution. This design ensures that differences in outcomes or behavior are attributable to differences in the agents themselves, rather than differences in underlying preferences.

This game has several features that make it well-suited for comparing agent types. First, its specific formulation is novel, providing a meaningful out-of-distribution domain for LLM agents.
Second, despite the controlled structure of the game, bargaining under partial information admits no Pareto-efficient, incentive-compatible solution \citep{MSImpossible1983}, forcing agents to navigate genuine tradeoffs between self-interest and cooperation. Third, we can vary strategic complexity by adjusting the number of chip types in players’ endowments. Finally, while the game is not itself incentive-compatible, it does admit a computable Pareto-optimal upper bound, which can benchmark performance.

We conduct an empirical study of how different agent types play this game under identical conditions. First, we run an incentivized online study with human participants ($N=216$).\footnote{Humans are paid in USD per their gains from trade beyond their initial endowment.} We then replicate the same randomized initial conditions in simulation with (i) LLM-based agents (\texttt{GPT-4o} and \texttt{Gemini 1.5 Pro}) and (ii) specialized Bayesian agents, enabling controlled comparisons across the three agent types. This setup enables an examination of two central research questions:

\begin{enumerate}[label=\textbf{RQ\arabic*:}, leftmargin=*]
\item \textbf{Performance:} How does the performance of LLM agents compare to bespoke statistical agents and humans?
\item \textbf{Process:} How do the trading strategies and resulting group dynamics of LLM agents compare to those of other agents?
\end{enumerate}

\begin{figure*}[ht!]
\centering
\includegraphics[width=.88\textwidth]{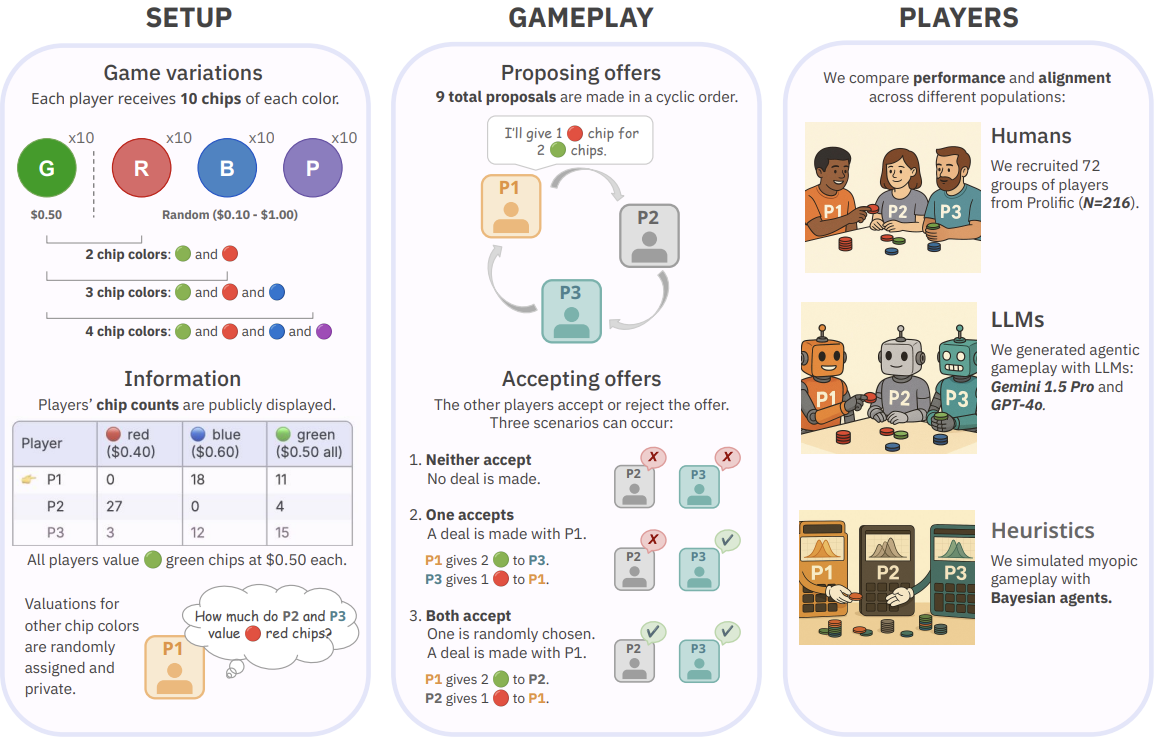}
\caption{Overview of the bargaining game. \textit{Left:} Game setup showing chip endowments and valuation structure. \textit{Center:} Gameplay mechanics for proposing and accepting offers. \textit{Right:} The three agent types evaluated in this study.}
\label{fig:overview}
\Description{Three-panel overview of a three-player bargaining game. Left: players receive chips of 2--4 colors; chip counts are public; green has a shared value, and other colors have private random values. Center: players propose in a fixed cycle; the two others accept or reject, yielding no deal, a bilateral deal, or a random tie-break if both accept. Right: agents include humans, LLMs (Gemini 1.5 Pro, GPT-4o), and heuristic baselines (myopic, Bayesian).}
\end{figure*}

We find that the game proves nontrivial: across all games and complexity levels, no agent type---human or artificial---achieves more than 80\% of the Pareto bound on average. The agents' performance varied substantially, and the processes by which they traded were remarkably different. Bayesian agents achieved the highest average surplus (up to 80\% of the Pareto bound) by proposing highly self-interested trades that were often rejected. LLM agents were capable of achieving similar surplus (62\%) to their human counterparts (59\%).\footnote{These statistics are for the medium-complexity game variation (3-chip). See Section~\ref{sec:context}.} 

Despite comparable outcomes, strategic behaviors between humans and GPT-4o agents diverged substantially. Humans offered trades that appeared more ``fair,'' while LLMs proposed more conservative trades that gave other LLMs within their group substantial surplus and were accepted at higher rates. Taken together, these results highlight the limitations of outcome-driven evaluation, as the similar surplus gains by humans and LLMs masked consequential differences in strategic behavior and procedural alignment.

In summary, this work makes three core contributions:
\begin{enumerate}[label=\textbf{\arabic*.}, leftmargin=*]
\item \textbf{A dynamic multi-agent environment for comparative evaluation.} We design and implement an open-source, incentivized bargaining game that enables ceteris paribus comparisons of behavior across agent types, with game variations of increasing complexity. The game is extensible to agent types not studied here (e.g., other LLMs). We also provide a linear programming formulation of the Pareto-optimal surplus as a theoretical benchmark for performance.

\item \textbf{An empirical sample of human, LLM, and Bayesian trading strategies.} We characterize how each agent type negotiates within a bargaining scenario under identical conditions: Bayesian agents aggressively maximize surplus and frequently reject offers; humans make strategic concessions motivated by fairness norms; LLMs negotiate conservatively and accept most offers.

\item \textbf{Evidence that outcome similarity can mask meaningful differences in behavior.} We show that aggregate outcomes (e.g., total surplus) can conceal substantial differences in strategies, interaction patterns, and normative alignment. This finding highlights the need to move beyond performance-based metrics for the responsible deployment of LLMs in high-stakes real-world settings.
\end{enumerate}

\section{Related Work}

Recent advances in LLMs' capabilities have dramatically broadened the scope of tasks that they can perform in scenarios involving collective reasoning, including understanding of social context, preference elicitation, strategic pricing, and natural-language negotiation \citep{fish2024algorithmic, soumalias2025llm, tessler2024ai, vaccaro2025negotiations, LubarsAIshould2019, meta2022human, kramar2022negotiation, deng2024llms, allouah2025buying}. 
LLMs have also demonstrated the capacity to perform economic tasks, replicating and predicting behaviors in bargaining, auctions, and other games \citep{aher2022using,argyle2022out,horton2023large,manning2024automated,manning2025general,xie2024canllm, zhu2024evidence, park2023generative, qian-etal-2025-mask}.

Yet demonstrating that LLMs \emph{can} perform economic tasks leaves open the question of \emph{how} they do so. \citet{imas2025agentic} study an experimental marketplace in which humans write prompts for AI agents that negotiate on their behalf. They find that, contrary to predictions of homogenized outcomes, agentic interactions exhibit greater dispersion than human-to-human negotiations. This suggests that prompts are not neutral instructions but carriers of human heterogeneity. 
Our work complements theirs: while they focus on how human characteristics propagate through delegation, we focus on comparing how different agent types behave when facing identical induced preferences, isolating strategic differences rather than principal heterogeneity.

These findings underscore a broader point: AI agents warrant behavioral study in their own right. 
\citet{rahwan2019machine} argue that the behavioral study of AI agents necessitates its own scientific research agenda. This is particularly critical in mixed-motive and multi-agent settings, where alignment is not static but dynamically negotiated. Recent work on virtual bargaining suggests that effective coordination requires agents to mentally simulate implicit social contracts rather than solely maximizing reward functions \citep{chater2023could}. Furthermore, \citet{bansal2019updates} highlight a ``performance-compatibility tradeoff'': updates that increase an agent's theoretical accuracy can paradoxically degrade team performance if they violate the user's mental model of the agent's error boundaries.

More generally, a broad theme in recent findings is that appropriate deployment depends not only on system capabilities \citep{jahani2025prompt}, but also on alignment with user expectations and social norms \citep{mozannar2023learning, Moehring2023Radio, Ziad2022Diagnose, palminteri2024, kapania2022, qian2024}. 
Addressing these alignment gaps requires evaluating agents in dynamic contexts with social consequences---a need recognized by recent calls for rigorous methods to steer and assess agent behavior \citep{raman2024steer, Goktas2025stratFoundation}. 
We contribute to this agenda by studying how humans, LLM-based agents, and Bayesian models behave in a dynamic bargaining task, offering a comparative lens on alignment in strategic social interactions.

\section{The Bargaining Game} \label{sec:context}

Figure~\ref{fig:overview} provides an overview of the game. 
Below, we describe the setup, information structure, and gameplay.

\paragraph{Setup.} 
A game consists of 3 players (any $n>2$ suffices) and $k$ distinct types of colored chips, where $k \in \{2, 3, 4\}$ varies across experimental conditions.\footnote{Game complexity is increasing in $k$. Proof provided in Appendix~\ref{app:increasing_complexity}.}
Players begin with 10 chips of each \textit{type}.
One chip type is always green, which serves as a shared numeraire publicly valued at \$0.50 by all players.
The remaining chip types (drawn from red, blue, and purple, depending on $k$) have private valuations: each player independently draws a value for each non-green chip type uniformly from [\$0.10, \$1.00].
While players do not know the valuations other players hold for non-green chips, they do know the distribution from which these values are drawn.
Crucially, this setup provides players with an incentive to trade because the possible mutual gains may be substantial.
For example, if one player values blue chips at \$0.20 and another values them at \$0.90, trading blue chips for green chips (the shared numeraire) allows both players to capture part of the \$0.70 valuation gap.

\paragraph{Gameplay.} Players take \textit{turns} in a fixed and initially randomized order, cycling through until each player has \textit{proposed} three times (9 total proposals for 3 players). 
On their turn, a player proposes a trade: offering chip(s) of one color in exchange for chip(s) of another. 
The non-proposing players simultaneously decide whether to accept or reject. 
If both accept, one is chosen at random to complete the trade.
Players may only propose or accept trades that they can fulfill with their current inventory. 
At the end of the game, each player's payout equals the difference in value between their final and initial chip holdings, floored at zero.
Throughout the game, players can see the number of chips each player holds, although not their opponents' valuations, along with a live history of all previous proposals and trades.

\section{A Pareto-Optimal Upper Bound}\label{sec:notation}

Next, we provide a theoretical upper bound analysis to benchmark agent performance in our empirical results. Due to the complexity of identifying a tractable, exact equilibrium in this setting, we instead compute a Pareto-optimal allocation via linear programming.

\subsection{Notation}

Consider a game $\textbf{M = (I, G, v, a)}$ with a set of agents $\textbf{I}$ and chips $\textbf{G}$. 
Each agent $i$ has a personal valuation $v_{ig}$ for each good $g$.
These valuations are static, independent across agents, and complete for each agent-good pair.
All agents have an initial allocation $a^0_{ig}$ of each chip, which indicates the amount of chip $g$ allocated to agent $i$. 
Given these allocations, an agent's \textbf{welfare} \textbf{$w_i$} is the sum of their valuations weighted by their allocations: $w_i = \sum_{g \in G} v_{ig}a_{ig}$. 
The total welfare $\textbf{w}$ is thus the sum of all individual welfare: $\textbf{w} = \sum_{i \in I} w_i= \sum_{i \in I} \sum_{g \in G} v_{ig}a_{ig}$.
An individual's \textbf{surplus} (and corresponding final payout) is defined as the difference between their final and initial welfare.

\subsection{Efficient Allocations}\label{sec:theory}

In our game, agents trade chips to improve their surplus. Given their initial endowments $a^0_{ig}$, mutually beneficial trades may exist that increase total welfare.
We provide an upper bound to the obtainable total welfare, assuming the Pareto condition that no player ends with worse utility than they began with.

\paragraph{Definition 1.}
\emph{An optimal allocation of chips} is any set of allocations $\mathbf{A^*} = \{a^*_{ig} \ | \ i \in I, g \in G\}$ that solves:
$$\argmax_{\mathbf{A}} \sum_{i \in I} \sum_{g \in G} v_{ig}a_{ig}$$
subject to:

\begin{enumerate}[label=(\roman*)]
\item \textbf{Conservation of goods}: $\sum_{i \in I} a_{ig} = \sum_{i \in I} a^0_{ig}, \forall \ g \in G$ 
\item \textbf{Pareto improvement}: $\sum_{g \in G} v_{ig}a_{ig} \ \geq \ \sum_{g \in G} v_{ig}a^0_{ig}, \forall \ i \in I$.
\item \textbf{Non-negativity}: $a_{ig} \geq 0 \forall \ i \in I, g \in G$.
\end{enumerate}

Under a centralized setting with full information, $\mathbf{A^*}$ represents a Pareto-efficient allocation that maximizes aggregate utility without making any agent worse off.
While this optimal allocation $\mathbf{A^*}$ may not be unique, the maximum total welfare $\mathbf{w^*}$ is unique.
This optimization problem can be solved efficiently using interior point methods. 
Its computational complexity is bounded by $O\bigl((|I||G|)^{3.5}\bigr)$. Note that this upper bound is unlikely to be achieved in practice, as no mechanism for decentralized bargaining with private information can guarantee both Pareto efficiency and incentive compatibility \cite{MSImpossible1983}.\footnote{See Appendix~\ref{app:no_dominant} for a proof that players have no dominant strategy in our game.}

\section[Experimental Setup]{Experimental Setup}\label{sec:setup}

We implemented the game using \emph{Deliberate Lab} \citep{deliberatelab}, an open-source platform for conducting multi-agent experiments. Our setup ensures full reproducibility, allowing for exact replication of game mechanics for benchmarking of future LLMs. Figures~\ref{fig:deliberatelabinterface} and ~\ref{app:dl_platform} provide visualizations of the game interface on Deliberate Lab.

In our experimental design, human participants were randomly assigned to groups of three to complete either the 2-chip, 3-chip, or 4-chip game. Humans also completed an end-of-task survey that asked them qualitative questions about their experience playing the game.\footnote{Full survey questions are provided in Appendix~\ref{app:survey}.}

These human game instances---defined by the unique initial chip endowments within a group---were then replicated in isolation using LLM and Bayesian agents. To facilitate a direct performance comparison, all groups remained homogeneous; no interactions occurred between different agent types (e.g., humans played exclusively against humans, and LLMs against LLMs).

\subsection{Human Subjects}\label{sec:human}

We recruited 216 U.S.-based participants from Prolific to complete the bargaining game on \textit{Deliberate Lab}.\footnote{Institutional IRB approval and informed consent were obtained; we applied no inclusion criteria for recruitment.} Each participant completed two games over a 30-minute session.
To incentivize play, one game was randomly selected for payment, and participants received their \textbf{surplus}---the difference between their final and initial chip values---as a bonus in addition to a \$4 base payment.
For example, a player whose chips increase in value from \$10 to \$14 earns a \$4 surplus bonus.
The average total payout was \$12.24 $\pm$ \$6.12. Each participant completed two games with different chip valuations and profiles. 
For each of the 2-, 3-, and 4-chip game variants, 24 groups of three players played 2 games each with 9 turns, yielding 1{,}296 trade proposals across 144 games.
Participants were not informed whether they were matched with the same partners across games. 
The two games are evaluated independently, as ordering and learning effects proved minimal.\footnote{Learning effects are discussed in Appendix~\ref{app:learning}.}

\subsection{LLM Agents}\label{sec:agents}

We replicated each human game with two types of LLM agents. 
First, we wanted to evaluate how LLMs perform ``out of the box,'' constructing a baseline agent using intentionally simple prompts adapted directly from the human instructions. Second, we wanted to see if we could improve performance by prompting the LLM to reason more deeply about the strategic context. 
To do this, we implemented a \textit{refined} agent that first generates multiple candidate proposals and then uses a second LLM call to select among them---mimicking the slower, more deliberate ``Type 2'' decision-making process described by \citet{kahneman2011thinking} and applied to LLMs by \citet{furniturewala-etal-2024-thinking}.
We tested both prompting strategies on OpenAI's \texttt{gpt-4o-2024-05-13} (\texttt{gpt-4o}, hereinafter) \citep{gpt4o} and Google's \texttt{gemini-1.5-pro-001} released on 2024-05-23 (\texttt{Gemini 1.5 Pro}) \citep{team2024gemini}, which were the latest full-capacity models from each family at the time of writing.\footnote{We also tested (\texttt{Gemini 2.5 Flash}, \texttt{GPT-o4-mini}), but they could not reliably propose valid trades. See Appendix~\ref{app:smaller_models} for further details.} 
Model temperature was set to 0.5.\footnote{A lower temperature risked agents repeatedly sampling a narrow subset of trades \citep{lambert2023alignment}. 
A higher temperature risked invalid trades \citep{arora2024optimizinglargelanguagemodel}.}
Prompts were designed using chain-of-thought \citep{wei2022chain}, and simulations were implemented using the open-source EDSL library \citep{horton2024edsl}.

Importantly, the replications were identical to the human games in terms of initial conditions and gameplay. 
At each turn, an LLM agent received the same information as its human counterpart, including private chip valuations, public chip holdings, and trade history. One agent proposed a trade per round, and the others evaluated it.

\begin{figure*}[t]
\centering
\includegraphics[width=.7\textwidth]{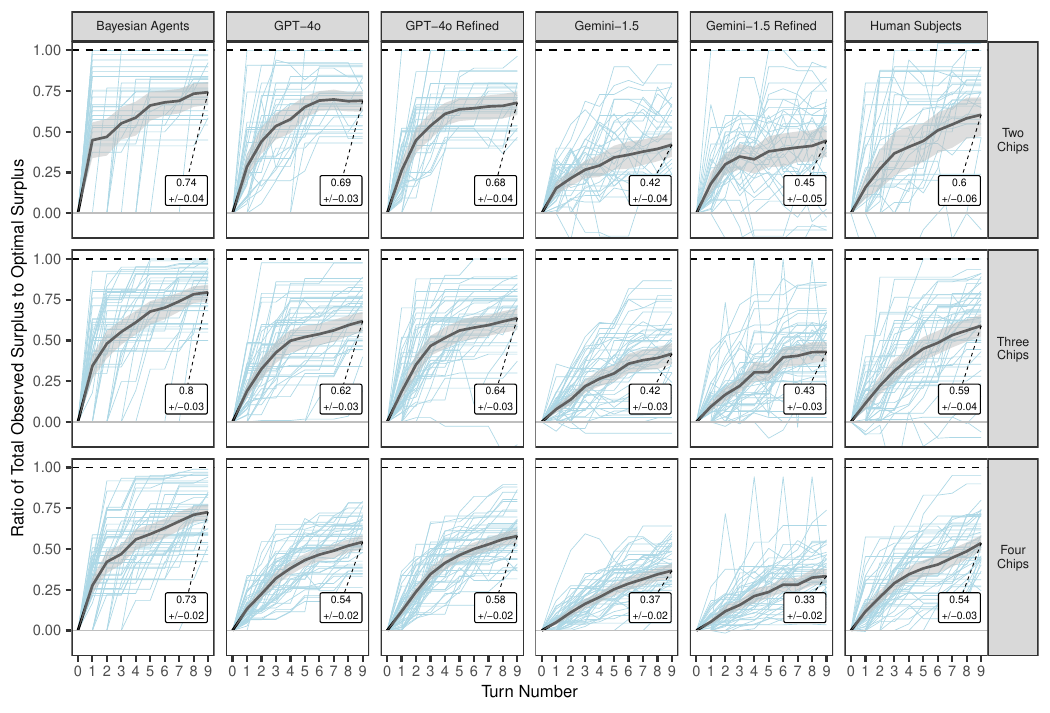}
\caption{Surplus trajectories across game complexities by agent type. Each blue line traces a game's surplus as a fraction of the Pareto-optimal benchmark; black lines show average trajectories with 95\% confidence bands.
The dashed line marks the theoretical ceiling (ratio = 1), and each panel displays the final average at Turn 9. Final surplus values are reported in Table~\ref{tab:summary_metric_table}.}
\Description{Surplus trajectories across game complexities by agent type. Each blue line traces a game's surplus as a fraction of the Pareto-optimal benchmark; black lines show average trajectories with 95\% confidence bands.
The dashed line marks the theoretical ceiling (ratio = 1), and each panel displays the final average at Turn 9.}
\label{fig:compare_agents_llm_2}
\end{figure*}

\subsection{Bayesian Agents}

We include a Bayesian agent as a task-specific benchmark---an explicitly programmed agent that optimizes within the known structure of the game.
We summarize the algorithm here and provide full details in Appendix~\ref{appendix:bayesian-algo}. 
Each agent $i$ forms a joint belief about the valuations of all other players, $ \mathbf{v}_{-i}$. When agent $i$ proposes a trade $(x_g, y_r)$
-- offering $x$ chips of type $g$ in exchange for $y$ chips of type $r$ -- they choose a trade to maximize their expected payoff:

\begin{figure*}[!ht]
\centering
\includegraphics[width=0.75\linewidth]{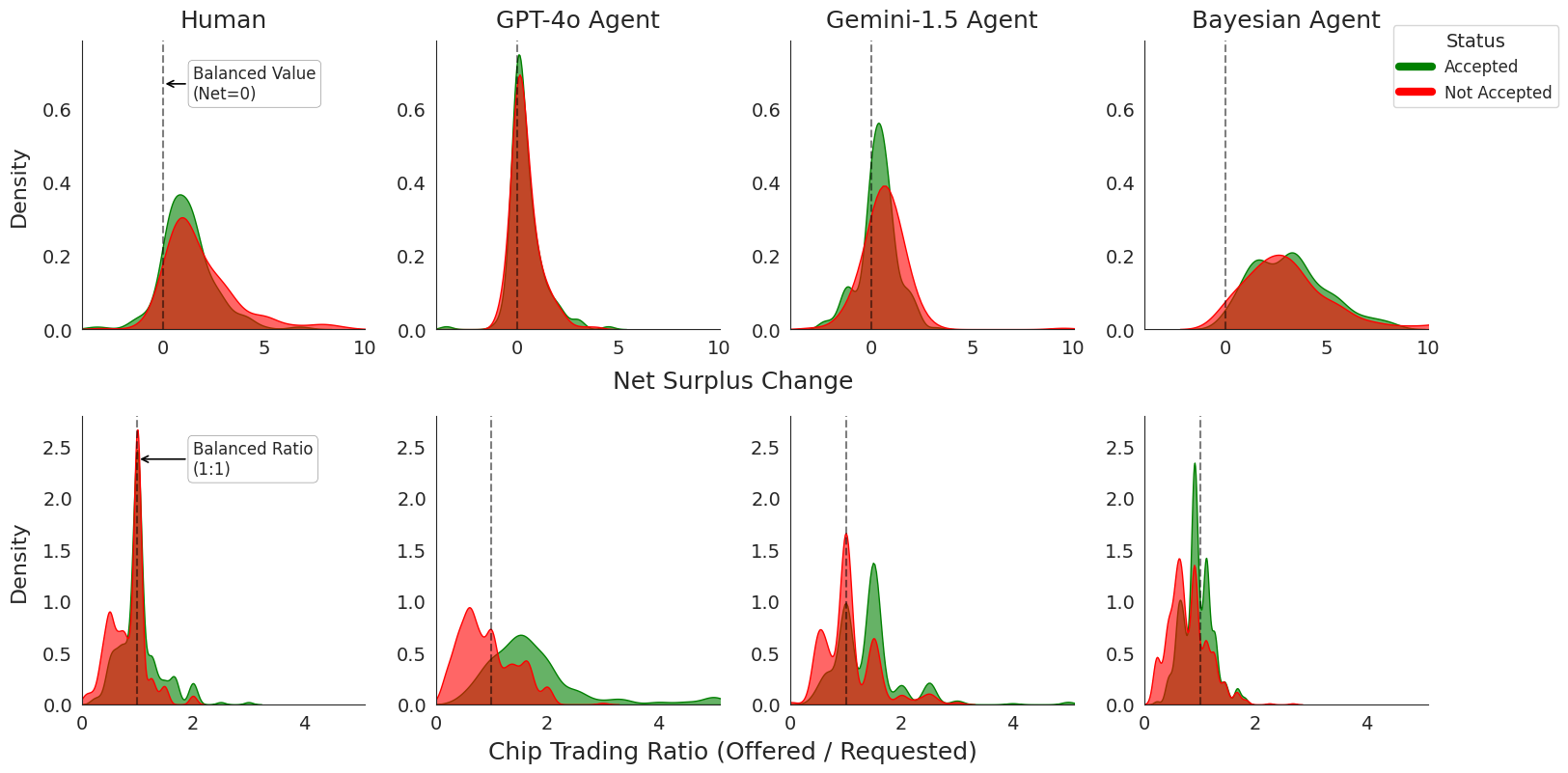}
\caption{Trading patterns in the 3-chip game. \textit{Top:} Net surplus change for the proposer. Values to the right of the dashed line represent \textit{positive surplus} to the proposer. \textit{Bottom:} Chip trading ratio (number of chips offered to requested). Accepted trades are shown in green, trades that were not accepted (for any reason) are in red. Values near the dashed line mark \textit{balanced exchange} (a 1:1 ratio of chips).}
\label{fig:trade-space}
\Description{Trading patterns in the 3-chip game. \textit{Top:} Net surplus change for the proposer. Values to the right of the dashed line represent \textit{positive surplus} to the proposer. \textit{Bottom:} Chip trading ratio (number of chips offered to requested). Accepted trades are shown in green, trades that were not accepted (for any reason) are in red. Values near the dashed line mark \textit{balanced exchange} (a 1:1 ratio of chips).}
\end{figure*}

\begin{equation*}
\max_{(\mathbf{x}, \mathbf{y})}
\sum_{\mathbf{v}_{-i}} 
\mathbf{1}\Bigl\{\exists\, j \ne i : \text{accept}\bigl(v_j, x_g, y_r\bigr)\Bigr\}
\;\times\;
\Delta u_i(v_i, x_g, y_r)
\;\times\;
B_{i}\bigl(\mathbf{v}_{-i}\bigr),
\end{equation*}
where $\text{accept}\bigl(v_j, x_g, y_r\bigr)$  indicates that agent $j$ with valuation $ v_j$  would accept the trade, $\Delta u_i(v_i, x_g, y_r) = v_{i,r}y_r-v_{i,g}x_g$ is agent $i$'s change in utility and $B_{i}(\mathbf{v}_{-i})$ is player $i$'s belief (probability) that the other agents' valuations are $\mathbf{v}_{-i}$. 
This agent assumes that their opponent's acceptance decision is \textit{myopically rational}, i.e., a receiver with valuation $v_j$ accepts if and only if they have sufficient chips for the trade and their expected utility is positive, i.e., $v_{j,g}x_g-v_{j,r}y_r > 0$.

All agents update their priors after observing acceptance decisions; they discard valuation states $\{v_j\}$ that contradict the observed event, assuming myopically rational opponents, and re-normalize the remaining probabilities.
When the trade is accepted, the proposer keeps only valuations that would be accepted in its distribution over the chosen receiver.
The receiver keeps only valuations that would be accepted in its distribution over the proposer. 
Bystanders update both priors accordingly. When there is no trade, all players discard valuations that conflict with the observed rejections.

\section{Results}\label{sec:results} 

We organize our findings around the two research questions.
For \textbf{RQ1} (performance), we compare aggregate surplus across agent types.
For \textbf{RQ2} (process), we examine trading patterns both at the turn- and game-level. We additionally report qualitative feedback from human participants on their strategies and experiences. 

\subsection{Performance} 

Figure~\ref{fig:compare_agents_llm_2} shows how total group surplus---the sum of all players' gains from trade---evolves over each game's nine turns. Bayesian agents consistently achieved the highest surplus across all game variants, reaching 74--80\% of the optimum.
\texttt{GPT-4o} performed comparably to human participants, while \texttt{Gemini 1.5 Pro} achieved lower aggregate surplus (42--45\% vs.\ 54--69\% for humans and \texttt{GPT-4o}).
The difference in performance between \texttt{GPT-4o} and the humans is negligible.
The refined prompting variants did not substantially improve performance for either LLM family.

Beyond the final surplus, several patterns stand out in these trajectories. 
Bayesian agents show notably concave paths—rapid early gains that flatten as obvious trades are exhausted.
In contrast, human and LLM trajectories rise more linearly, with \texttt{Gemini 1.5 Pro} showing the slowest average gains.
As game complexity increased from two to four chip types, all agent types captured a smaller share of the optimal surplus, though Bayesian agents remained the most robust to this increase.

\subsection{Procedural Alignment}\label{sec:alignment}

Given that humans and \texttt{GPT-4o} agents achieved comparable surplus, we now ask whether they arrived there through similar processes. In this section, we evaluate strategic differences between the human, \texttt{GPT-4o}, and Bayesian agents. We analyze behavior at two levels, since agents may accept short-term losses to position themselves for later gains, or conversely, capture early value at the cost of foreclosing better opportunities. We first examine the structure of individual proposals at the turn level—what each trade would yield if accepted, without considering future opportunities. We then evaluate decisions retrospectively—whether they were optimal given the opportunities that arose later in the game.

\subsubsection{Turn-level analysis}\label{sec:trade_patterns}

To characterize trading behavior at the turn level, we examine two features of each proposal: (i) the net surplus it would generate for the proposer, and (ii) the ratio of chips offered to chips requested.

Figure~\ref{fig:trade-space} shows these distributions for the 3-chip game.\footnote{2- and 4-chip results are qualitatively similar; figures provided in Appendix~\ref{app:trade_space}. Figure values are provided in Table~\ref{tab:summary_trade_table}.} The top row shows that humans generally proposed trades that would generate positive surplus, though some yield net-zero or negative returns. The bottom row shows that human proposals tend to be \textit{fair} in the number of chips exchanged. Both LLM families tend to propose trades that generate little surplus and more often result in a personal loss. On the chip trading ratio plot, a noticeable tail extends above the 1:1 line, indicating a tendency to offer more chips than requested (ratios up to 5:1).

By construction, Bayesian agents propose trades that give them surplus greater than zero---all density lies right of zero in the surplus plot.\footnote{Density mass below zero surplus for the Bayesian agents is an artifact of Gaussian smoothing. None of the Bayesian proposals generate a negative surplus by construction.} In the chip trading ratio row, Bayesian proposals cluster below 1:1 (ratio < 1); that is, they request more chips than they offer. Such proposals are frequently rejected.

These trading patterns reveal clear strategic differences.
Humans offer trades suggesting a consideration for balanced, fair exchanges \citep{danz2022belief}.
LLMs propose highly acceptable trades consistent with a ``concessionary posture''---offering many chips in return for a few. Bayesian agents adopt a value-extractive strategy \citep{fisher2011getting}, asking to receive more than they give, at the cost of frequent rejections. To evaluate whether any of these strategies were optimal ex-post, we next apply a regret-minimization framework. 

\begin{figure*}[ht!]
\centering
\includegraphics[width=0.8\linewidth]{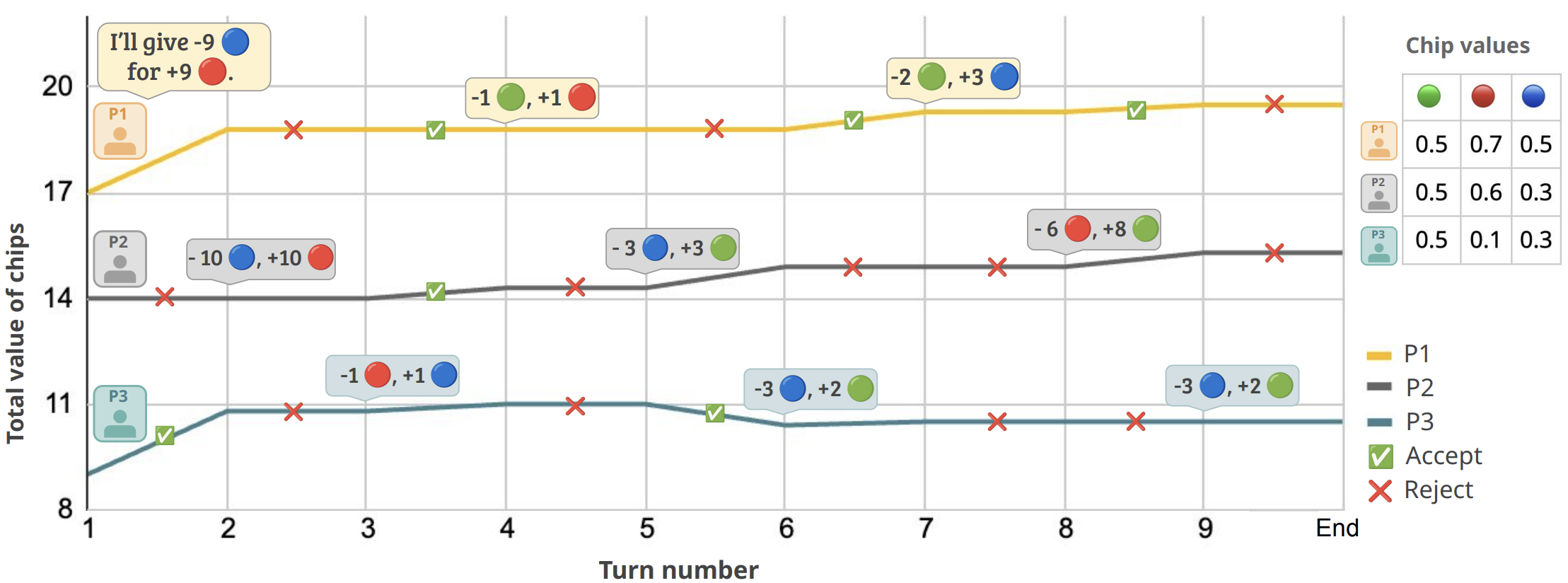}
\caption{Example of player chip value trajectories over nine turns of a trading game (3-chip). To avoid speculation, we do not analyze unobserved strategies. For example, we do not analyze Player 1's proposal on Turn 4 (an unaccepted trade) or Player 3's acceptance on Turn 5 (accepting a negative surplus offer).}
\label{fig:example-trajectory}
\Description{Example of player chip value trajectories over nine turns of a trading game (3-chip). To avoid speculation, we do not analyze unobserved strategies. For example, we do not analyze Player 1's proposal on Turn 4 (an unaccepted trade) or Player 3's acceptance on Turn 5 (accepting a negative surplus offer).}
\end{figure*}

\subsubsection{Retrospective analysis}
\label{sec:regret}

We now evaluate decisions retrospectively, asking whether they were optimal given the opportunities that actually materialized later in the game. At each turn, a player takes one of three actions: proposing, accepting, or declining a trade.
We classify each action by its regret properties:
\begin{itemize}
    \item \textbf{No regret}: There was no alternative move in a later round that would have yielded higher surplus.
    \item \textbf{Forced regret}: A higher-surplus option arose later, but earlier decisions (such as premature trades or overcommitting inventory) prevented the agent from acting on it.
    \item \textbf{Unforced regret}: A higher-surplus move was available to take (i.e., no budget or chip count constraints), yet the agent failed to take it.
\end{itemize}

We analyze how these regret classifications apply to the actions of the different agent types. To avoid speculative modeling of unobserved strategies, we restrict our analysis to a subset of trades where the agent's intent is clear. First, we only consider myopically rational transactions for proposers, acceptors, or decliners.\footnote{That is, only transactions yielding positive surplus. Note that Bayesian agents satisfy this by design.} 
Interpreting irrational trades would require assumptions we cannot credibly test, as they may be driven by factors such as reciprocity, signaling, or cognitive error. Second, when evaluating proposer behavior, we restrict our regret analysis to focus on accepted trades. The act of rejecting an offer could be a mistake, but it could also be strategic signaling. While acceptances can also stem from participant error, they provide observable realized gains that allow for objective regret calculation. 
Third, ``acceptors'' include those who intended to accept a trade, not only those who were selected to trade. Recall that this occurs when both other players in a game accept the trade; one is randomly chosen.

Figure~\ref{fig:example-trajectory} shows an example of a player's valuations as they trade over the course of a single game. On the left, private valuations are shown: Player 3 values blue chips at \$0.30 and red chips at \$0.10. On turn 1, Player 1 proposes to give 9 blue chips to receive 9 red chips in return. Player 3 is an acceptor and ends the turn with 19 blue chips and 1 red chip, netting a \$1.80 surplus. However, in turn 2, Player 2 offers a more lucrative offer: 10 blue chips for 10 red chips, which would give Player 3 a higher surplus of \$2.00 if they could accept it. However, Player 3 does not have enough red chips to clear the Turn 2 trade. Therefore, we characterize their action in Turn 1 as \textit{forced regret} by the acceptor. 

On turn 7, Player 2 makes an \textit{unforced regret} decline. Player 1 offers to give two green chips for three blue chips. Player 2 has enough chips to clear this trade at this point; if Player 2 were to accept, they would gain $0.5 * 2 - 0.3 * 3 = \$0.10$. However, they decline the profitable trade and do not recover the profits later in the game.\footnote{We do not classify proposals as \textit{unforced regret}; agents may send intentionally conservative offers to elicit signals on market dynamics.}

On turn 8, Player 2 makes a \textit{no regret proposal}. Player 2 offers to give six red chips to receive eight green chips, resulting in a surplus of $0.5 * 8 - 0.6 * 6 = \$0.40$. This was the best deal they could have made, as there were no better trade opportunities in the future. Additional details on the regret taxonomy are provided in Appendix~\ref{app:regret_description}.

Figure~\ref{fig:strategies} counts these actions across agent types and game complexity. Each panel corresponds to a game variant (2, 3, or 4 chips); rows group agents by role (proposer, acceptor, decliner).
Green bars indicate no-regret actions, blue bars unforced regret, and red bars forced regret. Because we restrict our analysis to unambiguous actions, these bars reflect the frequency of classifications within the filtered sample. The relative proportions of color within each bar, rather than the absolute bar length, provide the relevant comparison of strategic behavior across agent types.

\begin{figure*}[ht!]
\centering
\includegraphics[width=0.9\linewidth]{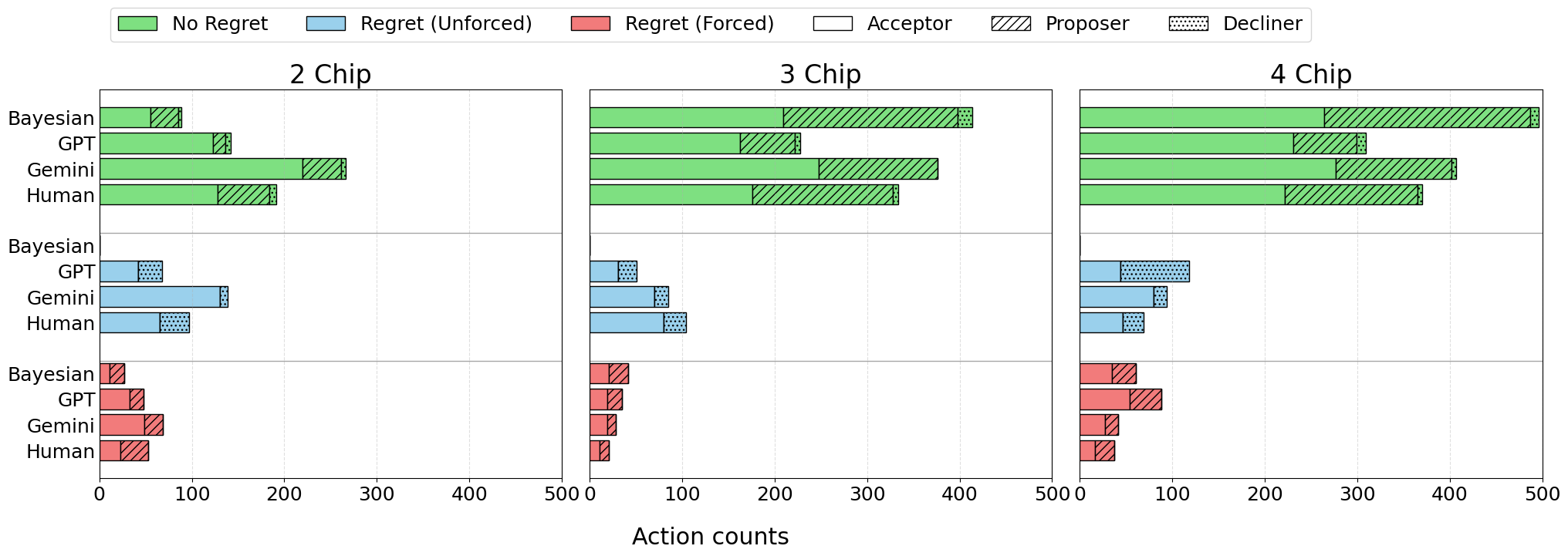}
\caption{Counts of actions by regret category across agent types and game complexity. Green is no regret, blue is unforced regret, and red is forced regret. Rows group agents by role (proposer, acceptor, decliner). Note: declining cannot cause forced regret since it does not commit chips.}
\label{fig:strategies}
\Description{Counts of actions by regret category across agent types and game complexity. Green is no regret, blue is unforced regret, and red is forced regret. Rows group agents by role (proposer, acceptor, decliner). Note: declining cannot cause forced regret since it does not commit chips.}
\end{figure*}

Several patterns emerge. The proportion of green (no regret) in Bayesian proposals relative to other classifications is high. That is, they consistently identify optimal trades.
In contrast, human and LLM proposers show substantial relative unforced regret (blue); more often than their Bayesian counterparts, they leave surplus on the table when better options were available. This is by construction; the Bayesian agents are myopically rational and accept all positive-surplus offers. Figure~\ref{fig:compare_agents_llm_2} shows that Bayesian agents achieve the highest realized surplus overall, suggesting that this myopic surplus-seeking policy may be well-suited for maximizing value in this environment.

\subsection{Qualitative Survey Feedback}

Finally, we examine participants' subjective experiences through post-game survey responses. Despite increasing game complexity, self-reported cognitive exertion, satisfaction, and performance certainty did not differ significantly across the 2-, 3-, and 4-chip conditions. This effort invariance is consistent with the use of fixed-complexity heuristics (such as the balanced 'equal-volume' trades observed in Figure~\ref{fig:trade-space}), which may allow the human players to maintain a stable cognitive load regardless of the underlying state-space complexity.

Participants additionally described their trading strategies in words.
Thematic analysis \citep{braun2006using} of these freeform responses revealed three primary themes: profit maximization (cited in 112 responses), collaborative strategy (56), and color optimization (55).\footnote{These distributions are visualized in Figure~\ref{fig:strategies2}. Responses were double-coded if they addressed multiple themes.} \textit{Profit maximization} was the most frequently cited approach (112 responses), with participants describing goals like maximizing their own chip gains. \textit{Collaborative} and \textit{color-focused} strategies appeared at similar rates (56 and 55 responses, respectively). Many participants prioritized fairness and mutual benefit, with explanations such as ``I aimed to be fair and reasonable with my trades,'' or ``I kept my offers low to encourage cautious players to engage.'' Others took a more aggressive approach, such as ``I traded aggressively early on to secure chips before others could deplete the supply.''

These qualitative accounts align with the observed human tendency toward reciprocal exchanges, where participants request a volume of chips similar to the amount offered. These accounts can also be interpreted alongside the self-reported effort invariance. For example, \textit{color-based} strategies can help players simplify decision-making and manage their cognitive load. Notably, however, none of these self-reported strategies (collaborative, profit-maximizing, or color-optimizing) correlated with actual game performance.

%%% Note: Checklist explicitly asks us to add this section.
\section{Discussion}\label{sec:discussion}

We briefly discuss behavioral patterns observed across agent types and offer recommendations for deployment and evaluation.

\subsection{Empirical Findings}

\paragraph{Human behaviors.} 
Our study sample comprised online strangers engaging in one-shot, mostly anonymous interactions.
There was effectively no reputational risk between players---they did not even know whether they were playing against the same humans in the second game.
Given this, a traditional game-theoretic model might predict more extractive behaviors prioritizing individual utility maximization \citep{rubinstein1982, binmore2007}. 
However, the humans offered trades consistent with fairness norms, i.e., the 1:1 ratio of chips offered to chips requested.
After the games, themes like coordination and fairness emerged as primary objectives. Such proposals generated less total surplus than Bayesian agents but were more frequently accepted, though humans still exhibited substantial unforced regret, leaving surplus on the table even when better trades were available.
In sum, these behaviors could suggest that humans anticipate that their human counterparts will reject unfair offers, and may be more individually rational than overall surplus would suggest.

These findings reveal a significant tension between traditional game-theoretic predictions and observed human behavior. Even in an environment with anonymity and one-shot interactions, specifically designed to minimize social accountability, human behavior remains conspicuously non-extractive. This also suggests that human objectives are not easily reduced to standard reward functions; rather, strategic reasoning appears deeply rooted in social context and background knowledge that non-human agents may fail to capture. In a mixed-agent future, it is unclear whether these human fairness norms will hold against non-human negotiators who may not share such norms.

\paragraph{LLM behaviors.}
Despite the novel domain, some \texttt{GPT-4o}-powered agents achieved human-level surplus with almost no additional prompting. They achieved this surplus through offering generally concessionary proposals ---giving more chips than they requested---which were readily accepted by other LLM agents. However, this acceptance-maximizing approach led to high rates of forced regret, as agents committed chips early and foreclosed superior opportunities later in the game.

While the underlying causes of this baseline conservative play style remain opaque, it likely arises from three intersecting factors: (i) instruction-tuning on cooperative, information-sharing dialogues where minimizing friction is implicitly rewarded \citep{wei2022chain}; (ii) a tendency toward risk-averse, passive responses exacerbated by a lack of outcome-driven feedback during the task \citep{ouyang2022training}; and (iii) a learned aversion to asymmetric outcomes, where generous offers serve as a heuristic to secure rapid coordination \citep{fisher2011getting}.

The implications of this default behavior are significant. In non-cooperative settings such as zero-sum negotiations, this overly concessionary tendency could lead LLMs to systematically over-compromise at baseline, in the absence of explicitly prompted behavioral changes. Unless post-training techniques evolve to explicitly account for strategic context, designers may need to intentionally steer these models to prevent such biases from being exploited by more extractive counterparts in competitive environments.

The performance gap between \texttt{GPT-4o} and \texttt{Gemini 1.5 Pro} may also reflect differences in model development cycles. \texttt{Gemini 1.5 Pro} was released approximately three months prior to \texttt{GPT-4o}, and the rapid pace of capability improvements in this period could account for some of the observed differences, though we cannot isolate model architecture from training data or other factors.

\paragraph{Bayesian behaviors.}
Despite their trades being frequently rejected, Bayesian agents consistently captured the highest surplus. Their proposals were predominantly no-regret, reflecting strategic precision that held even as game complexity increased.
While impressive, this strong performance (in terms of surplus) required us to write a bespoke statistical algorithm specifically aligned to the game’s incentives.
We have no reason to believe these behaviors will effectively translate to real-world negotiations, especially in repeated interactions requiring trust and reciprocity. Extending Bayesian agents to more socially compatible trading would require richer models of preferences, fairness, and adaptive behavior, potentially increasing complexity and decreasing robustness.

\subsection{Direct Implications}

Beyond the empirical results, a central contribution of this work is demonstrating the limits of outcome-based evaluation for agentic negotiation systems. 
Historically, models are trained and benchmarked using static, aggregate metrics such as total surplus. Had our analysis stopped at this level, one might reasonably conclude that some LLMs are functional substitutes for human negotiators.
However, we find substantial heterogeneity in how these LLMs behave relative to their human counterparts, both in their trading behavior and in the opportunities they later forego. While uncovering these procedural discrepancies in this study required novel evaluation methods that go beyond current, static agentic benchmarks, these procedural differences may be critical in a practitioner's decision to deploy agentic systems in a real-world scenario. 

Additionally, if one were considering deployment in this trading game, it is not clear whether LLMs or Bayesian agents are the better choice. For \textbf{satisficing} performance, we find that LLMs can generalize well to novel negotiation domains with minimal context-specific prompt-engineering.
Indeed, additional prompting provided no advantage over the baseline.
Both LLMs and Bayesian agents resolve negotiations near-instantaneously and are relatively costless at the margin compared to humans.
These efficiency advantages make automated agents appealing for high-volume or time-sensitive negotiations, even when beliefs about the other agents 
are imperfect.

For \textbf{optimal} performance, tradeoffs are more complex.
Human agents appear to trade according to ``unwritten'' fairness rules.
Bayesian agents maximize surplus but incur frequent rejections that could prove costly in repeated or reputation-sensitive settings.

Given the current model capabilities and limitations, an optimal implementation may involve a modular hybrid approach combining human values with the speed and reasoning of automated agents.
Here, we can imagine LLMs paired with planning or optimization components, such as Bayesian tools for valuation inference, inventory reasoning, or limited lookahead, to improve strategic foresight.
Humans as reviewers can then evaluate the final decisions to ensure they accord with the social context.

Concretely, we recommend that both researchers studying human–AI alignment and practitioners building or deploying AI negotiation agents in collective settings---especially those where agents operate alongside humans or act on their behalf---evaluate (i) \textit{outcome metrics}, such as deal rate and achieved surplus; (ii) \textit{procedural metrics}, such as strategic behaviors; and (iii) \textit{alignment metrics}, assessing whether an agent’s strategy reflects the intent, risk tolerance, and normative expectations of the human counterpart.

\subsection{Societal Implications}\label{sec:discussion-impacts}

Our findings also have broader societal implications for the deployment of AI agents in human-facing negotiation and coordination settings. In our study, humans consistently proposed fair trades despite no explicit incentives for doing so. 
One plausible explanation is that participants understood their counterparts to be other humans who value fairness, coordination, and mutual acceptance, and adjusted their behavior accordingly.
In this sense, human negotiation behavior appears to be shaped not only by material incentives but also by beliefs about the counterpart's identity and normative expectations.
This is consistent with much of the behavioral game theory literature \citep{camerer2003behavioral}.

These dynamics raise important questions for increasingly mixed-agent environments, where such beliefs may no longer be well calibrated.
If humans come to recognize that they are negotiating with concessionary LLMs, they may rationally adopt more aggressive strategies.
Conversely, repeated interaction with highly extractive, optimization-driven agents may undermine human fairness norms altogether, shifting negotiations toward more adversarial and less cooperative equilibria. 
In both cases, agents that achieve comparable surplus through fundamentally different negotiation styles may produce unexpected and differential outcomes in new settings.

\section{Limitations and Future Work}\label{sec:limitations}
Our study focuses on a single, stylized negotiation game with static valuations and single-shot interactions, conditions that align well with Bayesian agents’ task-specific optimization and may exaggerate their advantage.
While humans and LLMs bring general-purpose reasoning to the task, their broader capabilities in dynamic, multi-round, or reputation-driven negotiations remain unexplored.
Furthermore, our LLMs used minimal prompting, reflecting realistic deployment but limiting exploration of more strategic behaviors from alternative prompts or fine-tuning.

Beyond performance metrics, our analysis reveals distinct behavioral profiles but does not disentangle whether these arise from reasoning limitations, social norms, or inductive biases inherent to each system. Future work should investigate how humans perceive and trust AI negotiators with different behaviors. Another promising avenue for future work is to allow agents to communicate with natural language.
Finally, developing methods to steer LLM agents towards desired procedural norms represents a significant ongoing challenge and opportunity.
To support this agenda, we've released our game implementation on \emph{Deliberate Lab} \citep{deliberatelab}, enabling researchers to benchmark new models against the human, LLM, and Bayesian baselines established here as they emerge.

\newpage

\section{Conclusion}
As LLMs begin to match human-level outcomes in social decision-making tasks, the central question has shifted from \emph{whether} they can perform to \emph{how} they do so---and whether their strategies are robust, predictable, and normatively acceptable.

This study provides a controlled empirical approach for comparing humans, LLMs, and bespoke statistical agents in a dynamic bargaining setting. We organize our findings around two questions: how these agents \emph{perform} in terms of achieved surplus, and how they behave in the \emph{process} of getting there. For RQ1 (performance), Bayesian agents capture the most surplus, while \texttt{GPT-4o}-based agents can achieve human-level surplus. For RQ2 (process), we show that similar aggregate outcomes can arise from sharply different negotiation styles: humans tend toward fairness-constrained exchange, while LLMs default to concessionary, acceptance-seeking proposals.

These procedural differences are likely to matter more as LLM agents move beyond stylized games and static interactions into longer-horizon, higher-stakes interactions embedded in ongoing relationships. Agents that concede too readily invite exploitation, while agents that extract too aggressively risk rejection, mistrust, or norm violations. Because identifying these failure modes depends on strategic analysis rather than outcome comparison, static evaluations can be misleading. Responsible deployment, therefore, requires process-sensitive evaluation that makes strategic posture legible, quantifying missed opportunities and checking alignment with human stakeholders' goals, risk tolerance, and fairness norms.

\section{GenAI Usage Disclosure} 

LLMs were deployed as one of our populations of study.
Portions of Figure 1 were generated using \texttt{GPT-4o}. 
LLM-assisted tools were used for code editing in data analysis notebooks, proofreading, and light copyediting.
Generative AI was not used in other parts of the research process, including literature review, substantial data analysis, or writing.
The authors retain full responsibility for all content and conclusions presented herein.

\newpage

\begingroup
\small
\bibliographystyle{ACM-Reference-Format}
\bibliography{chip_chat}

@techreport{horton2023large,
  title={Large language models as simulated economic agents: What can we learn from homo silicus?},
  author={Horton, John J},
  year={2023},
  institution={National Bureau of Economic Research}
}

@article{danz2022belief,
  title={Belief elicitation and behavioral incentive compatibility},
  author={Danz, David and Vesterlund, Lise and Wilson, Alistair J},
  journal={American Economic Review},
  volume={112},
  number={9},
  pages={2851--2883},
  year={2022},
  publisher={American Economic Association 2014 Broadway, Suite 305, Nashville, TN 37203}
}

@inproceedings{xia-etal-2024-measuring,
    title = "Measuring Bargaining Abilities of {LLM}s: A Benchmark and A Buyer-Enhancement Method",
    author = "Xia, Tian  and
      He, Zhiwei  and
      Ren, Tong  and
      Miao, Yibo  and
      Zhang, Zhuosheng  and
      Yang, Yang  and
      Wang, Rui",
    editor = "Ku, Lun-Wei  and
      Martins, Andre  and
      Srikumar, Vivek",
    booktitle = "Findings of the Association for Computational Linguistics: ACL 2024",
    month = aug,
    year = "2024",
    address = "Bangkok, Thailand",
    publisher = "Association for Computational Linguistics",
    url = "https://aclanthology.org/2024.findings-acl.213/",
    doi = "10.18653/v1/2024.findings-acl.213",
    pages = "3579--3602"
}

@article{lambert2023alignment,
  title={The alignment ceiling: Objective mismatch in reinforcement learning from human feedback},
  author={Lambert, Nathan and Calandra, Roberto},
  journal={arXiv preprint arXiv:2311.00168},
  year={2023}
}

@book{fisher2011getting,
  title={Getting to yes: Negotiating agreement without giving in},
  author={Fisher, Roger and Ury, William L and Patton, Bruce},
  year={2011},
  publisher={Penguin}
}

@misc{gpt4o,
      title={GPT-4 Technical Report}, 
      author={OpenAI and Josh Achiam and Steven Adler and Sandhini Agarwal and Lama Ahmad and Ilge Akkaya and Florencia Leoni Aleman and Diogo Almeida and Janko Altenschmidt and Sam Altman and Shyamal Anadkat and Red Avila and Igor Babuschkin and Suchir Balaji and Valerie Balcom and Paul Baltescu and Haiming Bao and Mohammad Bavarian and Jeff Belgum and Irwan Bello and Jake Berdine and Gabriel Bernadett-Shapiro and Christopher Berner and Lenny Bogdonoff and Oleg Boiko and Madelaine Boyd and Anna-Luisa Brakman and Greg Brockman and Tim Brooks and Miles Brundage and Kevin Button and Trevor Cai and Rosie Campbell and Andrew Cann and Brittany Carey and Chelsea Carlson and Rory Carmichael and Brooke Chan and Che Chang and Fotis Chantzis and Derek Chen and Sully Chen and Ruby Chen and Jason Chen and Mark Chen and Ben Chess and Chester Cho and Casey Chu and Hyung Won Chung and Dave Cummings and Jeremiah Currier and Yunxing Dai and Cory Decareaux and Thomas Degry and Noah Deutsch and Damien Deville and Arka Dhar and David Dohan and Steve Dowling and Sheila Dunning and Adrien Ecoffet and Atty Eleti and Tyna Eloundou and David Farhi and Liam Fedus and Niko Felix and Simón Posada Fishman and Juston Forte and Isabella Fulford and Leo Gao and Elie Georges and Christian Gibson and Vik Goel and Tarun Gogineni and Gabriel Goh and Rapha Gontijo-Lopes and Jonathan Gordon and Morgan Grafstein and Scott Gray and Ryan Greene and Joshua Gross and Shixiang Shane Gu and Yufei Guo and Chris Hallacy and Jesse Han and Jeff Harris and Yuchen He and Mike Heaton and Johannes Heidecke and Chris Hesse and Alan Hickey and Wade Hickey and Peter Hoeschele and Brandon Houghton and Kenny Hsu and Shengli Hu and Xin Hu and Joost Huizinga and Shantanu Jain and Shawn Jain and Joanne Jang and Angela Jiang and Roger Jiang and Haozhun Jin and Denny Jin and Shino Jomoto and Billie Jonn and Heewoo Jun and Tomer Kaftan and Łukasz Kaiser and Ali Kamali and Ingmar Kanitscheider and Nitish Shirish Keskar and Tabarak Khan and Logan Kilpatrick and Jong Wook Kim and Christina Kim and Yongjik Kim and Jan Hendrik Kirchner and Jamie Kiros and Matt Knight and Daniel Kokotajlo and Łukasz Kondraciuk and Andrew Kondrich and Aris Konstantinidis and Kyle Kosic and Gretchen Krueger and Vishal Kuo and Michael Lampe and Ikai Lan and Teddy Lee and Jan Leike and Jade Leung and Daniel Levy and Chak Ming Li and Rachel Lim and Molly Lin and Stephanie Lin and Mateusz Litwin and Theresa Lopez and Ryan Lowe and Patricia Lue and Anna Makanju and Kim Malfacini and Sam Manning and Todor Markov and Yaniv Markovski and Bianca Martin and Katie Mayer and Andrew Mayne and Bob McGrew and Scott Mayer McKinney and Christine McLeavey and Paul McMillan and Jake McNeil and David Medina and Aalok Mehta and Jacob Menick and Luke Metz and Andrey Mishchenko and Pamela Mishkin and Vinnie Monaco and Evan Morikawa and Daniel Mossing and Tong Mu and Mira Murati and Oleg Murk and David Mély and Ashvin Nair and Reiichiro Nakano and Rajeev Nayak and Arvind Neelakantan and Richard Ngo and Hyeonwoo Noh and Long Ouyang and Cullen O'Keefe and Jakub Pachocki and Alex Paino and Joe Palermo and Ashley Pantuliano and Giambattista Parascandolo and Joel Parish and Emy Parparita and Alex Passos and Mikhail Pavlov and Andrew Peng and Adam Perelman and Filipe de Avila Belbute Peres and Michael Petrov and Henrique Ponde de Oliveira Pinto and Michael and Pokorny and Michelle Pokrass and Vitchyr H. Pong and Tolly Powell and Alethea Power and Boris Power and Elizabeth Proehl and Raul Puri and Alec Radford and Jack Rae and Aditya Ramesh and Cameron Raymond and Francis Real and Kendra Rimbach and Carl Ross and Bob Rotsted and Henri Roussez and Nick Ryder and Mario Saltarelli and Ted Sanders and Shibani Santurkar and Girish Sastry and Heather Schmidt and David Schnurr and John Schulman and Daniel Selsam and Kyla Sheppard and Toki Sherbakov and Jessica Shieh and Sarah Shoker and Pranav Shyam and Szymon Sidor and Eric Sigler and Maddie Simens and Jordan Sitkin and Katarina Slama and Ian Sohl and Benjamin Sokolowsky and Yang Song and Natalie Staudacher and Felipe Petroski Such and Natalie Summers and Ilya Sutskever and Jie Tang and Nikolas Tezak and Madeleine B. Thompson and Phil Tillet and Amin Tootoonchian and Elizabeth Tseng and Preston Tuggle and Nick Turley and Jerry Tworek and Juan Felipe Cerón Uribe and Andrea Vallone and Arun Vijayvergiya and Chelsea Voss and Carroll Wainwright and Justin Jay Wang and Alvin Wang and Ben Wang and Jonathan Ward and Jason Wei and CJ Weinmann and Akila Welihinda and Peter Welinder and Jiayi Weng and Lilian Weng and Matt Wiethoff and Dave Willner and Clemens Winter and Samuel Wolrich and Hannah Wong and Lauren Workman and Sherwin Wu and Jeff Wu and Michael Wu and Kai Xiao and Tao Xu and Sarah Yoo and Kevin Yu and Qiming Yuan and Wojciech Zaremba and Rowan Zellers and Chong Zhang and Marvin Zhang and Shengjia Zhao and Tianhao Zheng and Juntang Zhuang and William Zhuk and Barret Zoph},
      year={2024},
      eprint={2303.08774},
      archivePrefix={arXiv},
      primaryClass={cs.CL},
      url={https://arxiv.org/abs/2303.08774}, 
}

@inproceedings{park2023generative,
  title={Generative agents: Interactive simulacra of human behavior},
  author={Park, Joon Sung and O'Brien, Joseph and Cai, Carrie Jun and Morris, Meredith Ringel and Liang, Percy and Bernstein, Michael S},
  booktitle={Proceedings of the 36th annual acm symposium on user interface software and technology},
  pages={1--22},
  year={2023}
}

@book{kahneman2011thinking,
  title={Thinking, fast and slow},
  author={Kahneman, Daniel},
  year={2011},
  publisher={macmillan}
}

@unpublished{Goktas2025stratFoundation,
  TITLE = {{Strategic Foundation Models}},
  AUTHOR = {Goktas, Denizalp and Greenwald, Amy and Osogami, Takayuki and Patel, Roma and Leyton-Brown, Kevin and Schoenebeck, Grant and Cornelisse, Daphne and Daskalakis, Constantinos and Gemp, Ian and Horton, John and Parkes, David C and Pennock, David M. and Prakash, Arjun and Ravindranath, Sai Srivatsa and Smith, Max Olan and Swamy, Gokul and Vinitsky, Eugene and Wasserkrug, Segev and Wellman, Michael and Wu, Jibang and Xu, Haifeng and Zhang, Jiayao and Zhang, Yichi and Zhao, Sadie and Zhu, Quanyan},
  URL = {https://hal.science/hal-04925309},
  NOTE = {working paper or preprint},
  YEAR = {2025},
  MONTH = Feb,
  HAL_ID = {hal-04925309},
  HAL_VERSION = {v1},
}

@article{fish2024algorithmic,
  title={Algorithmic collusion by large language models},
  author={Fish, Sara and Gonczarowski, Yannai A and Shorrer, Ran I},
  journal={arXiv preprint arXiv:2404.00806},
  volume={7},
  year={2024}
}

@techreport{Moehring2023Radio,
 title = "Combining Human Expertise with Artificial Intelligence: Experimental Evidence from Radiology",
 author = "Agarwal, Nikhil and Moehring, Alex and Rajpurkar, Pranav and Salz, Tobias",
 institution = "National Bureau of Economic Research",
 type = "Working Paper",
 series = "Working Paper Series",
 number = "31422",
 year = "2023",
 month = "July",
 doi = {10.3386/w31422},
 URL = "http://www.nber.org/papers/w31422",
}

@techreport{horton2024edsl,
  title={EDSL: Expected Parrot Domain Specific Language for AI Powered Social Science},
  author={Horton, John and Filippas, Apostolos and Horton, Robin},
  year={2024},
  institution={Expected Parrot},
  type={Whitepaper}
}

@article{team2024gemini,
  title={Gemini 1.5: Unlocking multimodal understanding across millions of tokens of context},
  author={Team, Gemini and Georgiev, Petko and Lei, Ving Ian and Burnell, Ryan and Bai, Libin and Gulati, Anmol and Tanzer, Garrett and Vincent, Damien and Pan, Zhufeng and Wang, Shibo and others},
  journal={arXiv preprint arXiv:2403.05530},
  year={2024}
}

@inproceedings{LubarsAIshould2019,
 author = {Lubars, Brian and Tan, Chenhao},
 booktitle = {Advances in Neural Information Processing Systems},
 editor = {H. Wallach and H. Larochelle and A. Beygelzimer and F. d\textquotesingle Alch\'{e}-Buc and E. Fox and R. Garnett},
 pages = {},
 publisher = {Curran Associates, Inc.},
 title = {Ask not what AI can do, but what AI should do: Towards a framework of task delegability},
 url = {https://proceedings.neurips.cc/paper_files/paper/2019/file/d67d8ab4f4c10bf22aa353e27879133c-Paper.pdf},
 volume = {32},
 year = {2019}
}

@article{van2022walmart,
  title={How Walmart automated supplier negotiations},
  author={Van Hoek, Remko and DeWitt, Michael and Lacity, Mary and Johnson, Travis},
  journal={Harvard Business Review},
  volume={8},
  number={11},
  pages={2022},
  year={2022}
}

@article{gaarlandt2022shop,
  title={AI Agents Are Changing How People Shop. Here’s What That Means for Brands},
  author={Gaarlandt, Jur and Korver, Wesley and Furr, Nathan and  Shipilov, Andrew},
  journal={Harvard Business Review},
  year={2025}
}

@article{kramar2022negotiation,
  title={Negotiation and honesty in artificial intelligence methods for the board game of Diplomacy},
  author={Kram{\'a}r, J{\'a}nos and Eccles, Tom and Gemp, Ian and Tacchetti, Andrea and McKee, Kevin R and Malinowski, Mateusz and Graepel, Thore and Bachrach, Yoram},
  journal={Nature Communications},
  volume={13},
  number={1},
  pages={7214},
  year={2022},
  publisher={Nature Publishing Group UK London}
}

@misc{palminteri2024, title={The Moral Turing Test: Evaluating Human-LLM Alignment in Moral Decision-Making}, url={osf.io/preprints/psyarxiv/ct6rx_v1}, DOI={10.31234/osf.io/ct6rx}, publisher={PsyArXiv}, author={Palminteri, Stefano and Garcia, Basile and Qian, Crystal}, year={2024}, month={Oct}}

@inproceedings{kapania2022,
author = {Kapania, Shivani and Siy, Oliver and Clapper, Gabe and SP, Azhagu Meena and Sambasivan, Nithya},
title = {”Because AI is 100\% right and safe”: User Attitudes and Sources of AI Authority in India},
year = {2022},
isbn = {9781450391573},
publisher = {Association for Computing Machinery},
address = {New York, NY, USA},
url = {https://doi.org/10.1145/3491102.3517533},
doi = {10.1145/3491102.3517533},
booktitle = {Proceedings of the 2022 CHI Conference on Human Factors in Computing Systems},
articleno = {158},
numpages = {18},
location = {New Orleans, LA, USA},
series = {CHI '22}
}

@inproceedings{qian2024,
author = {Qian, Crystal and Wexler, James},
title = {Take It, Leave It, or Fix It: Measuring Productivity and Trust in Human-AI Collaboration},
year = {2024},
isbn = {9798400705083},
publisher = {Association for Computing Machinery},
address = {New York, NY, USA},
url = {https://doi.org/10.1145/3640543.3645198},
doi = {10.1145/3640543.3645198},
booktitle = {Proceedings of the 29th International Conference on Intelligent User Interfaces},
pages = {370–384},
numpages = {15},
location = {Greenville, SC, USA},
series = {IUI '24}
}

@article{wei2022chain,
  title={Chain-of-thought prompting elicits reasoning in large language models},
  author={Wei, Jason and Wang, Xuezhi and Schuurmans, Dale and Bosma, Maarten and Xia, Fei and Chi, Ed and Le, Quoc V and Zhou, Denny and others},
  journal={Advances in neural information processing systems},
  volume={35},
  pages={24824--24837},
  year={2022}
}

@article{braun2006using,
  title={Using thematic analysis in psychology},
  author={Braun, Virginia and Clarke, Victoria},
  journal={Qualitative research in psychology},
  volume={3},
  number={2},
  pages={77--101},
  year={2006},
  publisher={Taylor \& Francis}
}

@article{tessler2024ai,
  title={AI can help humans find common ground in democratic deliberation},
  author={Tessler, Michael Henry and Bakker, Michiel A and Jarrett, Daniel and Sheahan, Hannah and Chadwick, Martin J and Koster, Raphael and Evans, Georgina and Campbell-Gillingham, Lucy and Collins, Tantum and Parkes, David C and others},
  journal={Science},
  volume={386},
  number={6719},
  pages={eadq2852},
  year={2024},
  publisher={American Association for the Advancement of Science}
}

@article{meta2022human,
  title={Human-level play in the game of Diplomacy by combining language models with strategic reasoning},
  author={Bakhtin, Anton and Brown, Noam and Dinan, Emily and Farina, Gabriele and Flaherty, Colin and Fried, Daniel and Goff, Andrew and Gray, Jonathan and Hu, Hengyuan and others},
  journal={Science},
  volume={378},
  number={6624},
  pages={1067--1074},
  year={2022},
  publisher={American Association for the Advancement of Science}
}

@article{argyle2022out,
  title   = {Out of One, Many: Using Language Models to Simulate Human Samples},
  author  = {Argyle, Lisa P and Busby, Ethan C and Fulda, Nancy and Gubler, Joshua and Rytting, Christopher and Wingate, David},
  journal = {arXiv preprint arXiv:2209.06899},
  year    = {2022}
}

@article{Ziad2022Diagnose,
    author = {Mullainathan, Sendhil and Obermeyer, Ziad},
    title = {Diagnosing Physician Error: A Machine Learning Approach to Low-Value Health Care*},
    journal = {The Quarterly Journal of Economics},
    volume = {137},
    number = {2},
    pages = {679-727},
    year = {2021},
    month = {12},
    issn = {0033-5533},
    doi = {10.1093/qje/qjab046},
    url = {https://doi.org/10.1093/qje/qjab046},
    eprint = {https://academic.oup.com/qje/article-pdf/137/2/679/43336202/qjab046.pdf},
}

@misc{mozannar2023learning,
      title={Who Should Predict? Exact Algorithms For Learning to Defer to Humans}, 
      author={Hussein Mozannar and Hunter Lang and Dennis Wei and Prasanna Sattigeri and Subhro Das and David Sontag},
      year={2023},
      eprint={2301.06197},
      archivePrefix={arXiv},
      primaryClass={cs.LG},
      url={https://arxiv.org/abs/2301.06197}, 
}

@article{aher2022using,
  title   = {Using Large Language Models to Simulate Multiple Humans},
  author  = {Aher, Gati and Arriaga, Rosa I and Kalai, Adam Tauman},
  journal = {arXiv preprint arXiv:2208.10264},
  year    = {2022}
}

@techreport{manning2024automated,
  title={Automated social science: Language models as scientist and subjects},
  author={Manning, Benjamin S and Zhu, Kehang and Horton, John J},
  year={2024},
  institution={National Bureau of Economic Research}
}

@article{soumalias2025llm,
  title={LLM-Powered Preference Elicitation in Combinatorial Assignment},
  author={Soumalias, Ermis and Jiang, Yanchen and Zhu, Kehang and Curry, Michael and Seuken, Sven and Parkes, David C},
  journal={arXiv preprint arXiv:2502.10308},
  year={2025}
}

@misc{vaccaro2025negotiations,
      title={Advancing AI Negotiations: New Theory and Evidence from a Large-Scale Autonomous Negotiations Competition}, 
      author={Michelle Vaccaro and Michael Caoson and Harang Ju and Sinan Aral and Jared R. Curhan},
      year={2025},
      eprint={2503.06416},
      archivePrefix={arXiv},
      primaryClass={cs.AI},
      url={https://arxiv.org/abs/2503.06416}, 
}

@article{Corvin2024alibaba,
  title={Web Summit 2024: Alibaba conjures up global trade AI, Portugal launches own LLM},
  author={Corvin, Ann-Marie},
  journal={TechInformed},
  year={2024},
  month={November},
  day={13},
  url={https://www.techinformed.com/web-summit-2024-alibaba-conjures-up-global-trade-ai-portugal-launches-own-llm/}
}

@article{PYMNTS2025openai,
  title={OpenAI Turns ChatGPT Into Shopping Assistant With New Commerce Features},
  author={{PYMNTS}},
  journal={PYMNTS},
  year={2025},
  month={April},
  day={28},
  url={https://www.pymnts.com/news/artificial-intelligence/2025/openai-turns-chatgpt-into-shopping-assistant-with-new-commerce-features/}
}

@article{MSImpossible1983,
title = {Efficient mechanisms for bilateral trading},
journal = {Journal of Economic Theory},
volume = {29},
number = {2},
pages = {265-281},
year = {1983},
issn = {0022-0531},
doi = {https://doi.org/10.1016/0022-0531(83)90048-0},
url = {https://www.sciencedirect.com/science/article/pii/0022053183900480},
author = {Roger B Myerson and Mark A Satterthwaite}
}

@inproceedings{
      xie2024canllm,
      title={Can Large Language Model Agents Simulate Human Trust Behavior?},
      author={Chengxing Xie and Canyu Chen and Feiran Jia and Ziyu Ye and Shiyang Lai and Kai Shu and Jindong Gu and Adel Bibi and Ziniu Hu and David Jurgens and James Evans and Philip Torr and Bernard Ghanem and Guohao Li},
      booktitle={The Thirty-eighth Annual Conference on Neural Information Processing Systems},
      year={2024},
      url={https://openreview.net/forum?id=CeOwahuQic}
      }

@misc{vafa2024general,
      title={Do Large Language Models Perform the Way People Expect? Measuring the Human Generalization Function}, 
      author={Keyon Vafa and Ashesh Rambachan and Sendhil Mullainathan},
      year={2024},
      eprint={2406.01382},
      archivePrefix={arXiv},
      primaryClass={cs.CL},
      url={https://arxiv.org/abs/2406.01382}, 
}

@misc{jahani2025prompt,
      title={Prompt Adaptation as a Dynamic Complement in Generative AI Systems}, 
      author={Eaman Jahani and Benjamin S. Manning and Joe Zhang and Hong-Yi TuYe and Mohammed Alsobay and Christos Nicolaides and Siddharth Suri and David Holtz},
      year={2025},
      eprint={2407.14333},
      archivePrefix={arXiv},
      primaryClass={cs.HC},
      url={https://arxiv.org/abs/2407.14333}, 
}

@inproceedings{brown2020language,
  title={Language Models are Few-Shot Learners},
  author={Brown, Tom and Mann, Benjamin and Ryder, Nick and Subbiah, Melanie and Kaplan, Jared D and Dhariwal, Prafulla and Neelakantan, Arvind and Shyam, Pranav and Sastry, Girish and Askell, Amanda and Agarwal, Sandhini and Herbert-Voss, Ariel and Krueger, Gretchen and Henighan, Tom and Child, Rewon and Ramesh, Aditya and Ziegler, Daniel and Wu, Jeffrey and Winter, Clemens and Hesse, Chris and Chen, Mark and Sigler, Eric and Litwin, Mateusz and Gray, Scott and Chess, Benjamin and Clark, Jack and Berner, Christopher and McCandlish, Sam and Radford, Alec and Sutskever, Ilya and Amodei, Dario},
  booktitle={Advances in Neural Information Processing Systems},
  volume={33},
  year={2020}
}

@article{smith1976induced,
 ISSN = {00028282},
 URL = {http://www.jstor.org/stable/1817233},
 author = {Vernon L. Smith},
 journal = {The American Economic Review},
 number = {2},
 pages = {274--279},
 publisher = {American Economic Association},
 title = {Experimental Economics: Induced Value Theory},
 urldate = {2025-12-25},
 volume = {66},
 year = {1976}
}

@article{imas2025agentic,
  title={Agentic Interactions},
  author={Imas, Alex and Lee, Kevin and Misra, Sanjog},
  journal={SSRN Electronic Journal},
  year={2025},
  month={December},
  doi={10.2139/ssrn.5875162},
  url={https://ssrn.com/abstract=5875162}
}

@inproceedings{hendrycks2021mmlu,
  title={Measuring massive multitask language understanding},
  author={Hendrycks, Dan and Burns, Collin and Basart, Steven and Zou, Andy and Mazeika, Mantas and Song, Dawn and Steinhardt, Jacob},
  booktitle={International Conference on Learning Representations},
  year={2021}
}

@article{GPTsAreGPTs2024,
author = {Tyna Eloundou  and Sam Manning  and Pamela Mishkin  and Daniel Rock },
title = {GPTs are GPTs: Labor market impact potential of LLMs},
journal = {Science},
volume = {384},
number = {6702},
pages = {1306-1308},
year = {2024},
doi = {10.1126/science.adj0998},
URL = {https://www.science.org/doi/abs/10.1126/science.adj0998},
eprint = {https://www.science.org/doi/pdf/10.1126/science.adj0998},
}

@article{brynjolfsson2023generative,
    author = {Brynjolfsson, Erik and Li, Danielle and Raymond, Lindsey},
    title = {Generative AI at Work*},
    journal = {The Quarterly Journal of Economics},
    volume = {140},
    number = {2},
    pages = {889-942},
    year = {2025},
    month = {02},
    issn = {0033-5533},
    doi = {10.1093/qje/qjae044},
    url = {https://doi.org/10.1093/qje/qjae044},
    eprint = {https://academic.oup.com/qje/article-pdf/140/2/889/61701561/qjae044.pdf},
}

@incollection{Shahidi2025Coasean,
 title = "The Coasean Singularity? Demand, Supply, and Market Design with AI Agents",
 author = "Peyman Shahidi and Gili Rusak and Benjamin S. Manning and Andrey Fradkin and John J. Horton",
 BookTitle = "The Economics of Transformative AI",
 Publisher = "University of Chicago Press",
 year = "2025",
 month = "December",
 URL = "http://www.nber.org/chapters/c15309",
}

@misc{manning2025general,
      title={General Social Agents}, 
      author={Benjamin S. Manning and John J. Horton},
      year={2025},
      eprint={2508.17407},
      archivePrefix={arXiv},
      primaryClass={econ.GN},
      url={https://arxiv.org/abs/2508.17407}, 
}

@inproceedings{ouyang2022training,
  title={Training Language Models to Follow Instructions with Human Feedback},
  author={Ouyang, Long and Wu, Jeff and Jiang, Xu and Almeida, Diogo and Wainwright, Carroll L. and Mishkin, Pamela and Zhang, Chong and Agarwal, Sandhini and Slama, Katarina and Ray, Alex and Schulman, John and Hilton, Jacob and Kelton, Fraser and Miller, Luke and Simens, Maddie and Askell, Amanda and Welinder, Peter and Christiano, Paul and Leike, Jan and Lowe, Ryan},
  booktitle={Advances in Neural Information Processing Systems},
  year={2022}
}

@misc{allouah2025buying,
title={What Is Your AI Agent Buying? Evaluation, Implications and Emerging Questions for Agentic E-Commerce}, 
author={Amine Allouah and Omar Besbes and Josué D Figueroa and Yash Kanoria and Akshit Kumar},
year={2025},
eprint={2508.02630},
archivePrefix={arXiv},
primaryClass={cs.AI},
url={https://arxiv.org/abs/2508.02630}, 
}

@inproceedings{qian-etal-2025-mask,
    title = "To Mask or to Mirror: Human-{AI} Alignment in Collective Reasoning",
    author = {Qian, Crystal  and
      Parisi, Aaron T  and
      Bouleau, Cl{\'e}mentine  and
      Tsai, Vivian  and
      Lebreton, Ma{\"e}l  and
      Dixon, Lucas},
    editor = "Christodoulopoulos, Christos  and
      Chakraborty, Tanmoy  and
      Rose, Carolyn  and
      Peng, Violet",
    booktitle = "Proceedings of the 2025 Conference on Empirical Methods in Natural Language Processing",
    month = nov,
    year = "2025",
    address = "Suzhou, China",
    publisher = "Association for Computational Linguistics",
    url = "https://aclanthology.org/2025.emnlp-main.122/",
    doi = "10.18653/v1/2025.emnlp-main.122",
    pages = "2398--2423",
    ISBN = "979-8-89176-332-6"
}

@ARTICLE{binmore2007,
title = {Ken Binmore, Playing for Real: A Text on Game Theory, Oxford University Press (2007) ISBN 978-0-19-530057-4 639 pp},
author = {Young, H.},
year = {2007},
journal = {Games and Economic Behavior},
volume = {59},
number = {2},
pages = {411-412},
url = {https://EconPapers.repec.org/RePEc:eee:gamebe:v:59:y:2007:i:2:p:411-412}
}

@Article{rubinstein1982,
journal={Econometrica},
author={Rubinstein, Ariel},
title={Perfect Equilibrium in a Bargaining Model},
year={1982},
month={January},
pages={97-109},
volume={50},
number={1},
doi={None},
url={https://ideas.repec.org/a/ecm/emetrp/v50y1982i1p97-109.html},
}

@misc{arora2024optimizinglargelanguagemodel,
      title={Optimizing Large Language Model Hyperparameters for Code Generation}, 
      author={Chetan Arora and Ahnaf Ibn Sayeed and Sherlock Licorish and Fanyu Wang and Christoph Treude},
      year={2024},
      eprint={2408.10577},
      archivePrefix={arXiv},
      primaryClass={cs.SE},
      url={https://arxiv.org/abs/2408.10577}, 
}

@inproceedings{
deng2024llms,
title={{LLM}s at the Bargaining Table},
author={Yuan Deng and Vahab Mirrokni and Renato Paes Leme and Hanrui Zhang and Song Zuo},
booktitle={Agentic Markets Workshop at ICML 2024},
year={2024},
url={https://openreview.net/forum?id=n0RmqncQbU}
}

@inproceedings{zhu2024evidence,
title={Evidence from the Synthetic Laboratory: Language Models as Auction Participants},
author={Kehang Zhu and John Joseph Horton and Yanchen Jiang and David C. Parkes and Anand V. Shah},
booktitle={NeurIPS 2024 Workshop on Behavioral Machine Learning},
year={2024},
url={https://openreview.net/forum?id=FB9mTtJpJI}
}

@inproceedings{furniturewala-etal-2024-thinking,
    title = "{\textquotedblleft}Thinking{\textquotedblright} Fair and Slow: On the Efficacy of Structured Prompts for Debiasing Language Models",
    author = "Furniturewala, Shaz  and
      Jandial, Surgan  and
      Java, Abhinav  and
      Banerjee, Pragyan  and
      Shahid, Simra  and
      Bhatia, Sumit  and
      Jaidka, Kokil",
    editor = "Al-Onaizan, Yaser  and
      Bansal, Mohit  and
      Chen, Yun-Nung",
    booktitle = "Proceedings of the 2024 Conference on Empirical Methods in Natural Language Processing",
    month = nov,
    year = "2024",
    address = "Miami, Florida, USA",
    publisher = "Association for Computational Linguistics",
    url = "https://aclanthology.org/2024.emnlp-main.13/",
    doi = "10.18653/v1/2024.emnlp-main.13",
    pages = "213--227"
}

@misc{deliberatelab,
      title={Deliberate Lab: A Platform for Real-Time Human-AI Social Experiments}, 
      author={Crystal Qian and Vivian Tsai and Michael Behr and Nada Hussein and Léo Laugier and Nithum Thain and Lucas Dixon},
      year={2025},
      eprint={2510.13011},
      archivePrefix={arXiv},
      primaryClass={cs.HC},
      url={https://arxiv.org/abs/2510.13011}, 
}

@article{raman2024steer,
  title={STEER: Assessing the economic rationality of large language models},
  author={Raman, Narun and Lundy, Taylor and Amouyal, Samuel and Levine, Yoav and Leyton-Brown, Kevin and Tennenholtz, Moshe},
  journal={arXiv preprint arXiv:2402.09552},
  year={2024}
}

@book{camerer2003behavioral,
  title={Behavioral game theory: Experiments in strategic interaction},
  author={Camerer, Colin},
  year={2003},
  publisher={Princeton university press}
}

@article{rahwan2019machine,
  title={Machine behaviour},
  author={Rahwan, Iyad and Cebrian, Manuel and Obradovich, Nick and Bongard, Josh and Bonnefon, Jean-Fran{\c{c}}ois and Breazeal, Cynthia and Crandall, Jacob W and Christakis, Nicholas A and Couzin, Iain D and Jackson, Matthew O and others},
  journal={Nature},
  volume={568},
  number={7753},
  pages={477--486},
  year={2019},
  publisher={Nature Publishing Group UK London}
}

@inproceedings{bansal2019updates,
  title={Updates in human-ai teams: Understanding and addressing the performance/compatibility tradeoff},
  author={Bansal, Gagan and Nushi, Besmira and Kamar, Ece and Weld, Daniel S and Lasecki, Walter S and Horvitz, Eric},
  booktitle={Proceedings of the AAAI conference on artificial intelligence},
  volume={33},
  pages={2429--2437},
  year={2019}
}

@article{chater2023could,
  title={How could we make a social robot? A virtual bargaining approach},
  author={Chater, Nick},
  journal={Philosophical Transactions of the Royal Society A},
  volume={381},
  number={2251},
  pages={20220040},
  year={2023},
  publisher={The Royal Society}
}
\endgroup

\clearpage
\appendix

\section{Algorithms and Proofs} \label{app:algos_proofs}

\subsection{Empirical Computation of Increasing Game Complexity}\label{app:increasing_complexity}
\paragraph{Claim.} \emph{The decision space increases quadratically as the number of chip colors increases.}

\paragraph{Proof.} This is a simple consequence that in games with $k$ different colors of chips, there are $\binom{k}{2}$ possible pairs of chips to transact. Since the distribution of opponents' valuations is symmetric across chips (except the green numeraire chip), each proposer must consider $O(k^2)$ different feasible trades. Indeed, if we empirically compute the expected number of myopically rational trades\footnote{We define a myopically rational trade to be one where the proposer gains surplus and assigns some probability that at least one opponent would gain surplus by accepting the trade.} in the starting configuration of the 2, 3, and 4-chip game, we arrive at 37.1, 120.7, 250.7 respectively. That is, the 3-chip game has approximately $\binom{3}{2}$ as many myopically rational trades in expectation as the 2-chip game, and the 4-chip game has approximately $\binom{4}{2}$ times as many. This approximation becomes more precise as the number of chips increases, since the numeraire's relative impact is reduced.

\subsection{Proof of No Dominant Strategy}\label{app:no_dominant}
\paragraph{Claim.}
\emph{No agent \(i \in \mathbf{I}\) in the game \(\mathcal{M}\) has a dominant strategy.}

\paragraph{Setup.}
Let \(\mathbf{I}\) be the set of agents, \(\mathbf{G}\) be the set of goods. 
For each agent \(i \in \mathbf{I}\) and good \(g \in \mathbf{G}\), there is a valuation \(v_{ig}\in \mathbb{R}_{\ge 0}\).
The (feasible) allocations are given by \(a_{ig}\), the amount of good \(g\) allocated to agent \(i\).

The \textbf{welfare} of agent \(i\) is  
\[
w_i \;=\; \sum_{g \in \mathbf{G}} \, v_{ig} \, a_{ig}.
\]

The \textbf{total social welfare} is  
\[
w \;=\; \sum_{i \in \mathbf{I}} \, w_i \;=\; \sum_{i \in \mathbf{I}} \sum_{g \in \mathbf{G}} v_{ig} \, a_{ig}.
\]

Let each agent \(i\) have a strategy space \(\mathcal{S}_i\). A (pure) strategy \(s_i \in \mathcal{S}_i\) is a complete description of agent \(i\)’s behavior in the market (e.g., how they propose, reject, accept, etc.).
We write \(\mathbf{s}_{-i}\) for a strategy profile of all agents other than \(i\). 
Let \(u_i(s_i, \mathbf{s}_{-i})\) denote agent \(i\)’s resulting payoff (or final utility/welfare) under strategy \(s_i\) against opponents’ strategies \(\mathbf{s}_{-i}\). 
A strategy \(s_i^*\) is called \textit{dominant} for agent \(i\) if, for \textbf{all} \(\mathbf{s}_{-i}\in \mathcal{S}_{-i}\), 
\[
u_i\bigl(s_i^*, \mathbf{s}_{-i}\bigr)
\;\;\ge\;\;
u_i\bigl(s_i, \mathbf{s}_{-i}\bigr)
\quad
\text{for all } s_i \in \mathcal{S}_i.
\]
In other words, \(s_i^*\) guarantees the highest possible utility for \(i\), irrespective of what other players do.

\paragraph{Proof.}

% \textbf{Proof:}

Assume, by way of contradiction, that there exists a dominant strategy \(s_i^*\) for some agent \(i\). By the definition of dominance, \(s_i^*\) must satisfy
\[
u_i\bigl(s_i^*, \mathbf{s}_{-i}\bigr)
\;\ge\;
u_i\bigl(s_i, \mathbf{s}_{-i}\bigr)
\quad\text{for all } s_i \in \mathcal{S}_i
\text{ and for all } \mathbf{s}_{-i} \in \mathcal{S}_{-i}.
\]

Construct two different scenarios of other agents’ strategies and valuations, denoted by \(\mathbf{s}_{-i}^A\) and \(\mathbf{s}_{-i}^B\). These scenarios can differ in (i) how other agents value the goods \(\{v_{jg}: j \neq i\}\), and (ii) the demands or offers they make (i.e.\ how \(\mathbf{s}_{-i}\) translates into allocations \(a_{jg}\)).

\begin{itemize}
\item In scenario \(A\), suppose the other agents are willing to trade or allocate a certain good \(g^*\) to agent \(i\) only if \(i\) offers a high price or concedes in some manner. A specialized alternative strategy \(s_i^A\) (instead of \(s_i^*\)) might yield a strictly higher payoff if \(i\) trades aggressively for \(g^*\). 
\item In scenario \(B\), suppose the others behave differently (e.g., they no longer place much value on \(g^*\) but highly value some other good). Now, a different specialized strategy \(s_i^B\) might yield a strictly higher payoff (because the trading strategy for \(g^*\) in scenario \(A\) is no longer beneficial here).
\end{itemize}

\textbf{Contradiction of dominance.}
Since \(A\) and \(B\) are feasible opponent profiles, no single \(s_i^*\) can weakly dominate
both tailored responses \(s_i^A\) and \(s_i^B\) across them. In particular,
\begin{align}
\exists\,\mathbf{s}_{-i}^A \;:\;&
u_i\!\left(s_i^A,\mathbf{s}_{-i}^A\right) \;>\; u_i\!\left(s_i^*,\mathbf{s}_{-i}^A\right),
\\
\exists\,\mathbf{s}_{-i}^B \;:\;&
u_i\!\left(s_i^B,\mathbf{s}_{-i}^B\right) \;>\; u_i\!\left(s_i^*,\mathbf{s}_{-i}^B\right).
\end{align}
This contradicts the definition of \(s_i^*\) as a dominant strategy.

Because the above argument does not rely on any special assumption beyond the existence of at least two different plausible profiles \(\mathbf{s}_{-i}^A\) and \(\mathbf{s}_{-i}^B\), we conclude that no strategy for agent \(i\) can be dominant. Since \(i\) was arbitrary, no agent in the game has a dominant strategy. \(\quad \square\)

\clearpage

\subsection{Multi-Agent Trading Algorithm with Bayesian Learning} \label{appendix:bayesian-algo}

\begin{algorithm}[h]
\small 
\caption{Multi-agent Trading Algorithm with Bayesian Learning} 

\begin{algorithmic}[1]
\Statex \textbf{Inputs:}
\Statex \quad $\mathbf{I}$: set of players
\Statex \quad $\mathbf{G}$: set of goods
\Statex \quad $\mathbf{A^0}$: initial allocation $\{a_{ig}\}$ for all $i \in \mathbf{I}, g \in \mathbf{G}$
\Statex \quad $\{B_i(\mathbf{v}_{-i})\}_{i\in \mathbf{I}}$: each player's prior belief over other players' valuations
\vspace{.5em}
\State $\mathbf{A} \gets \mathbf{A^0}$ 
\Comment{Initialize current allocation}
\ForAll{$i \in \mathbf{I}$}
\State \textbf{initialize} $B_i(\mathbf{v}_{-i})$ 
\Comment{Each agent's prior over other agents' valuations}
\EndFor 
\State $possible\_trades \gets \texttt{true}$
\While{$possible\_trades$}
\State $trades\_found \gets \texttt{false}$
\State $I_{\text{shuffled}} \gets \text{RandomShuffle}(\mathbf{I})$
\For{$k = 1$ \textbf{to} $|\mathbf{I_{\text{shuffled}}}|$}
\For{$l = k+1$ \textbf{to} $|\mathbf{I_{\text{shuffled}}}|$}
\State $(i, j) \gets (I_{\text{shuffled}}[k], I_{\text{shuffled}}[l])$

\State 
\begin{varwidth}[t]{\linewidth}
$(\mathbf{A^*_{ij}},\, \textit{outcome},\, r) \gets 
\textsc{SolveA}^*\Bigl(i,j,\mathbf{G},\{a_{ig}, a_{jg}\}_{g \in \mathbf{G}},B_i,B_j\Bigr)$
\end{varwidth}
\Comment{Proposer $i$ finds best trade; $r$ is the responding agent (who accepts/rejects)}

\If{$\textit{outcome} = \textsc{accept}$}
\Comment{Trade executed}
\State $\{a_{ig},\, a_{jg}\}_{g \in \mathbf{G}} \gets \mathbf{A^*_{ij}}$
\State $trades\_found \gets \texttt{true}$
\EndIf

\State $\textsc{BayesianUpdate}\bigl(i,\, j,\, r,\, \textit{outcome},\, B_i,\, B_j,\, \{B_k\}_{k \neq i,j}\bigr)$
\Comment{All players update beliefs based on accept/reject}
\EndFor
\EndFor
\State $possible\_trades \gets trades\_found$
\EndWhile
\State \Return $\mathbf{A}$
\end{algorithmic}
\end{algorithm}

\noindent \textbf{Explanation of key subroutines:}
\begin{itemize}
\item \(\textsc{SolveA}^*(i, j, \dots)\):
\begin{itemize}
\item \textbf{Proposer’s optimization.} 
Agent \(i\) (the proposer) solves 
\begin{equation}
\begin{aligned}
\max_{(\mathbf{x}, \mathbf{y})}\ 
&\sum_{\mathbf{v}_{-i}}
\mathbf{1}\!\left\{
\exists r\neq i:\ 
\textsc{accept}\bigl(v_r;\mathbf{x},\mathbf{y}\bigr)
\right\}
\\
&\qquad\qquad\times 
\bigl(
u_i(h_i-\mathbf{x}+\mathbf{y}) - u_i(h_i)
\bigr)
\times B_i(\mathbf{v}_{-i}).
\end{aligned}
\end{equation}
where \(\mathbf{x}\) are the chips given up by \(i\) and \(\mathbf{y}\) are the chips requested, \(h_i\) denotes \(i\)’s current holdings, and \(B_i\) is \(i\)’s belief over others’ valuations.
\item \textbf{Receiver’s acceptance criterion.}
A potential responder \(r\) accepts the trade \((\mathbf{x}, \mathbf{y})\) \emph{myopically} if 
\[
u_r\bigl(h_r - \mathbf{y} + \mathbf{x};\,v_r\bigr) 
\;>\; 
u_r\bigl(h_r;\,v_r\bigr).
\]
The subroutine returns \(\mathbf{A^*_{ij}}\) (the updated allocation for \(i\) and \(j\)), the \(\textit{outcome}\) (\textsc{accept} or \textsc{reject}), and the identity of the actual \emph{responder} \(r\) (in case multiple receivers are considered, or if $j$ is always the designated responder, $r = j$).
\end{itemize}

\item \(\textsc{BayesianUpdate}\bigl(i, j, r, \textit{outcome}, B_i, B_j, \{B_k\}_{k \neq i,j}\bigr)\):
\begin{itemize}
\item On \textsc{acceptance}: 
All agents discard any valuation states inconsistent with $r$ accepting \((\mathbf{x}, \mathbf{y})\). 
Specifically, 
\[
v_r \text{ is retained only if } u_r(h_r - \mathbf{y} + \mathbf{x}; v_r) > u_r(h_r; v_r). 
\]
\item On \textsc{rejection}:
All agents discard any valuation states for $r$ that would have \emph{accepted} \((\mathbf{x}, \mathbf{y})\), since that contradicts the observed rejection. 
\item Then, each agent renormalizes its belief distributions \(B_k\) to sum to 1 over the remaining (still-plausible) valuations.
\end{itemize}
\end{itemize}

\noindent This procedure repeats until no further beneficial trades can be found. The Bayesian updates ensure that, over time, agents refine their beliefs about each other’s valuations based on observed accept/reject decisions, thereby \emph{learning} which trades are more likely to succeed.

\section{Human Data Collection}\label{app:human_dc}

\subsection{Implementation on Deliberate Lab}\label{app:dl_platform}

\begin{figure*}
\centering
\includegraphics[width=.9\textwidth]{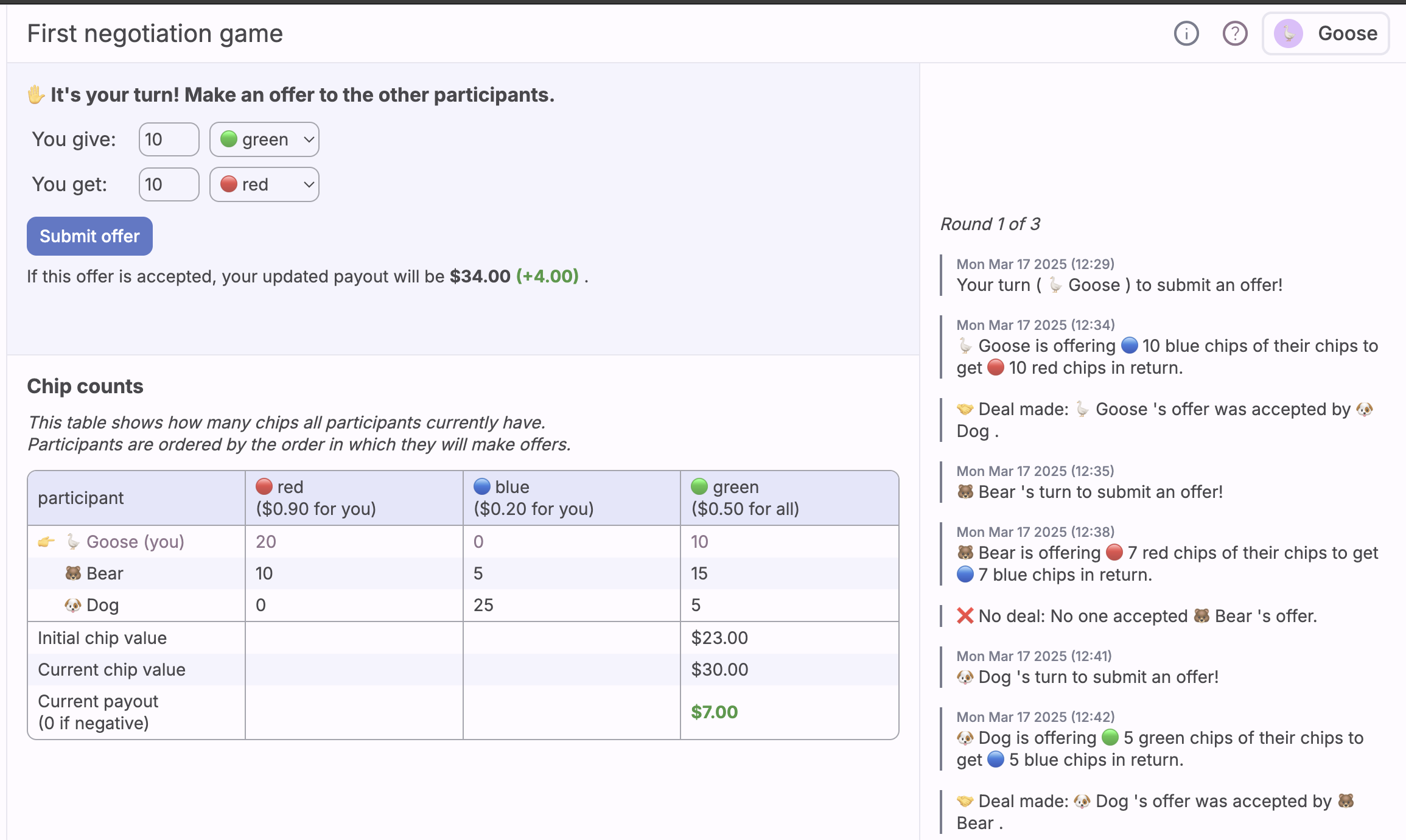}
\caption{The game interface exactly as it was shown to participants. The top left indicates that they are currently the \textit{proposer}, offering an interface to propose an offer. A table below displays all players' chip volumes. To the right, the public ledger shows a history of game rounds, turns, and trade statuses.}
\label{fig:deliberatelabinterface}
\Description{The game interface exactly as it was shown to participants. The top left indicates that they are currently the \textit{proposer}, offering an interface to propose an offer. A table below displays all players' chip volumes. To the right, the public ledger shows a history of game rounds, turns, and trade statuses.}
\end{figure*}

\paragraph{Participant interface.} Upon entering Deliberate Lab through a web link, participants proceed through a multi-stage experiment including informed consent, game instructions, comprehension checks, and payout information. Upon completing the final comprehension check, they wait in a ``Lobby'' stage for other participants. When three participants are in the lobby, they are sent an invitation to join a live bargaining game (Figure~\ref{fig:deliberatelabinterface}). Following the game, there is a post-game survey. For anonymity, we used a Deliberate Lab feature that assigns participants an anonymous animal avatar (e.g., ``Bear'') as they join the experiment.

\begin{figure*}[ht!]
\centering
\includegraphics[width=\textwidth]{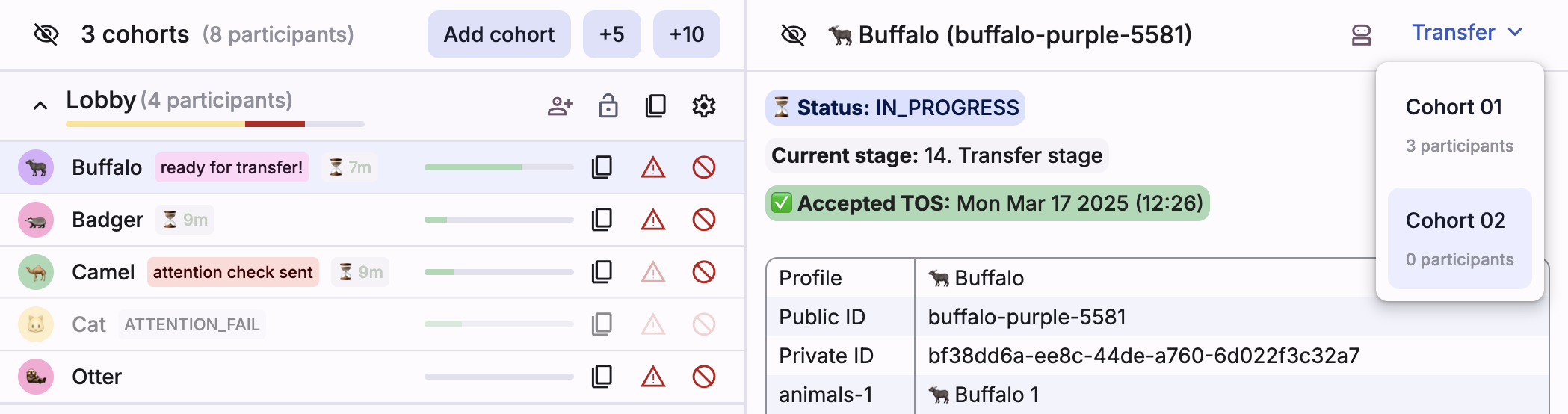}
\caption{Experiment dashboard for managing participants. On the left, experimenters can view participants as they enter the experiment, send attention checks, and/or remove participants. On the right, experimenters can monitor the status of a selected participant and assign them to a cohort of three for gameplay.}
\Description{Experiment dashboard for managing participants. On the left, experimenters can view participants as they enter the experiment, send attention checks, and/or remove participants. On the right, experimenters can monitor the status of a selected participant and assign them to a cohort of three for gameplay.}
\label{app:deliberatelabdashboard}
\end{figure*}

\begin{figure}[ht!]
\centering
\includegraphics[width=0.75\linewidth]{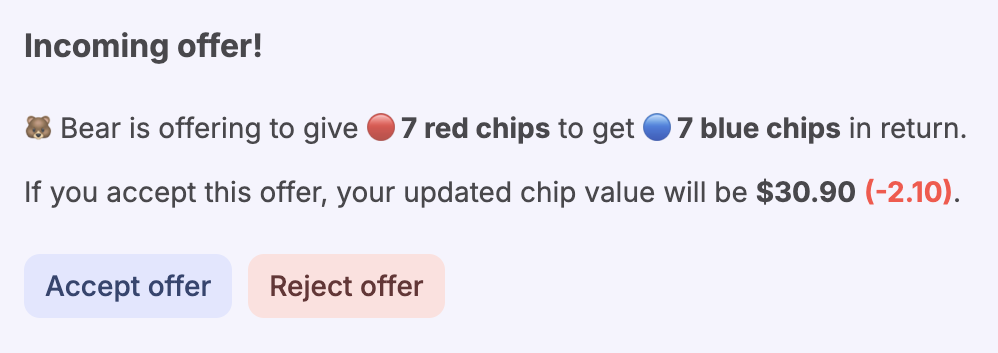}
\vspace{0.6em}
\includegraphics[width=0.75\linewidth]{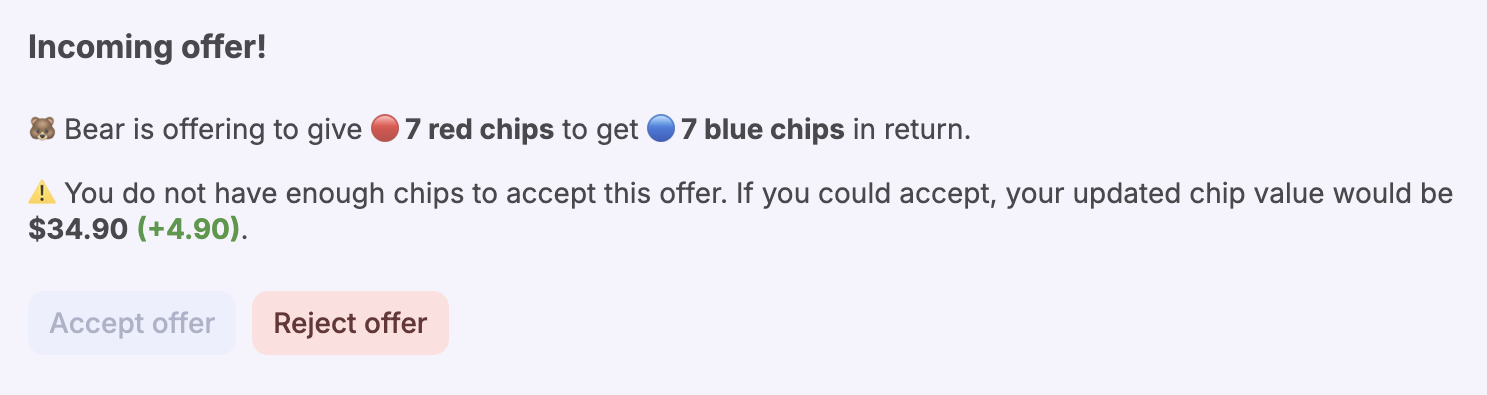}
\caption{Accepting or rejecting a trade. The bottom image shows a disabled ``accept'' option.}
\label{app:deliberatelabresponder}
\Description{Accepting or rejecting a trade. The bottom image shows a disabled ``accept'' option.}
\end{figure}

\paragraph{Bargaining game user interface.}
To submit trade proposals, participants used text and selection fields to specify the number and type of chip, respectively. The interface automatically calculated and displayed the projected total chip value if the proposal were to be accepted (Figure~\ref{fig:deliberatelabinterface}); it also prevented players from submitting invalid offers by disabling the ``submit'' button. For accepting or rejecting proposals, the interface calculated and displayed the projected total chip value if the player were to accept the proposal. It also disabled acceptance for offers the player could not fulfill (Figure~\ref{app:deliberatelabresponder}, bottom). All participants were shown a table of all players' chip quantities and a log of actions (i.e., when trades were proposed, accepted, and rejected, and when each round and turn of the game changed). Turn order was determined randomly.

\paragraph{Experimenter interface.}\label{ref:experimenter} To conduct experiments, the experimenter used Deliberate Lab's dashboard (Figure~\ref{app:deliberatelabdashboard}) to monitor participants as they progressed through the experiment's information and comprehension check stages, then manually transferred participants into groups of three to play the bargaining game once the player was ready to be transferred. The experimenter was also able to send attention checks to participants (e.g., if several minutes elapsed without any visible progress) and remove them from the experiment (e.g., if attention checks went unanswered).

\section{Post-Game Participant Survey}\label{app:survey}

\begin{table}
  \centering
  \footnotesize
  \begin{tabular}{lcccc}
    \toprule
    Game variation & Competitive & Certainty & Satisfaction & Exertion \\
    \midrule
    \textit{2-chip} & 5.57 (0.34) & 5.57 (0.33) & 6.68 (0.31) & 7.11 (0.24) \\
    \textit{3-chip} & 5.32 (0.34) & 5.92 (0.33) & 7.07 (0.30) & 7.10 (0.25) \\
    \textit{4-chip} & 5.75 (0.34) & 6.04 (0.33) & 7.46 (0.26) & 7.46 (0.19) \\
    \bottomrule
  \end{tabular}  \caption{Summary statistics of self-assessment Likert variables (mean and standard error); full survey questions in Table~\ref{tab:post_game_survey_questions}. There was no significant change in certainty or exertion as game complexity increased.}
    \label{tab:summary-self-assessment}
\end{table}

\begin{figure}[ht!]
\centering
\includegraphics[width=.7\linewidth]{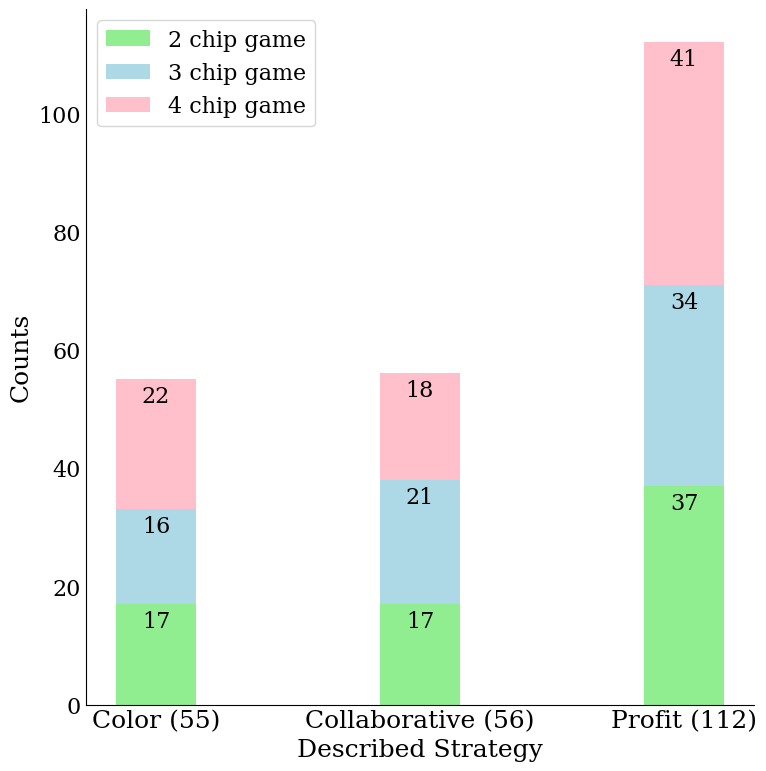}
\caption{ Distribution of self-reported strategies across game complexity (thematic analysis). Participants described their trading approach as focused on profit maximization, collaborative play, or color optimization. Profit-focused strategies were most common (112 responses), while collaborative (56) and color-focused (55) strategies appeared at similar rates. The distribution across 2-, 3-, and 4-chip games was relatively consistent within each strategy type.}
\label{fig:strategies2}
\Description{ Distribution of self-reported strategies across game complexity (thematic analysis). Participants described their trading approach as focused on profit maximization, collaborative play, or color optimization. Profit-focused strategies were most common (112 responses), while collaborative (56) and color-focused (55) strategies appeared at similar rates. The distribution across 2-, 3-, and 4-chip games was relatively consistent within each strategy type.}
\end{figure}

The survey questions are provided in Table~\ref{tab:post_game_survey_questions}. Responses to Likert-scale questions are aggregated in Table~\ref{tab:summary-self-assessment}. Qualitative thematic analysis is provided in Figure~\ref{fig:strategies2}.

\begin{table*}[ht!]
\footnotesize
\centering
\caption{Post-game survey questions.}

\label{tab:post_game_survey_questions}
\begin{tabular}{p{.3cm} p{9cm} p{4cm}}
\hline
& Survey question & Response format \\
\hline
1 & Please describe your strategy in the game in a few sentences. & Freeform text \\
\addlinespace
2 & Please describe any experiences or background that may have influenced your performance in the game. & Freeform text \\
\addlinespace
3 & On a scale from 1 to 10, how would you rate your trading strategy, where 1 is highly \textbf{competitive} (focused mainly on your own gains) and 10 is highly \textbf{collaborative} (focused on other players' potential gains and mutual benefits) & \makecell[tl]{1 through 10 scale;\\ 1: ``highly competitive'',\\ 10: ``highly collaborative''} \\
\addlinespace
4 & On a scale from 1 to 10, how certain are you that you made the best possible trades during the experiment? & \makecell[tl]{1 through 10 scale;\\ 1: ``Not at all confident'',\\ 10: ``Very confident''}  \\
\addlinespace
5 & On a scale from 1 to 10, how satisfied are you with your final trading outcomes? & \makecell[tl]{1 through 10 scale;\\ 1: ``Not at all satisfied'',\\ 10: ``Very satisfied''}  \\
\addlinespace
6 & On a scale of 1 to 10, how would you rate the mental effort you put into today's trading games, where 1 means 'I barely engaged in any thinking' and 10 means 'I put in significant mental effort'? & \makecell[tl]{1 through 10 scale;\\ 1: ``No mental exertion at all'',\\ 10: ``Very high mental exertion''}  \\
\addlinespace
7 & Please provide any additional context on your answers above. & Freeform text \\
\addlinespace
8 & Please help us to improve this experiment. How was your experience today? Were there any elements of the instructions or gameplay that you found confusing? & Freeform text \\
\hline
\end{tabular}
\end{table*}

\newpage \clearpage 

\section{Additional Clarification on Strategic Regret Classifications}\label{app:regret_description}

\begin{figure*}[ht!]
\centering
\includegraphics[width=\linewidth]{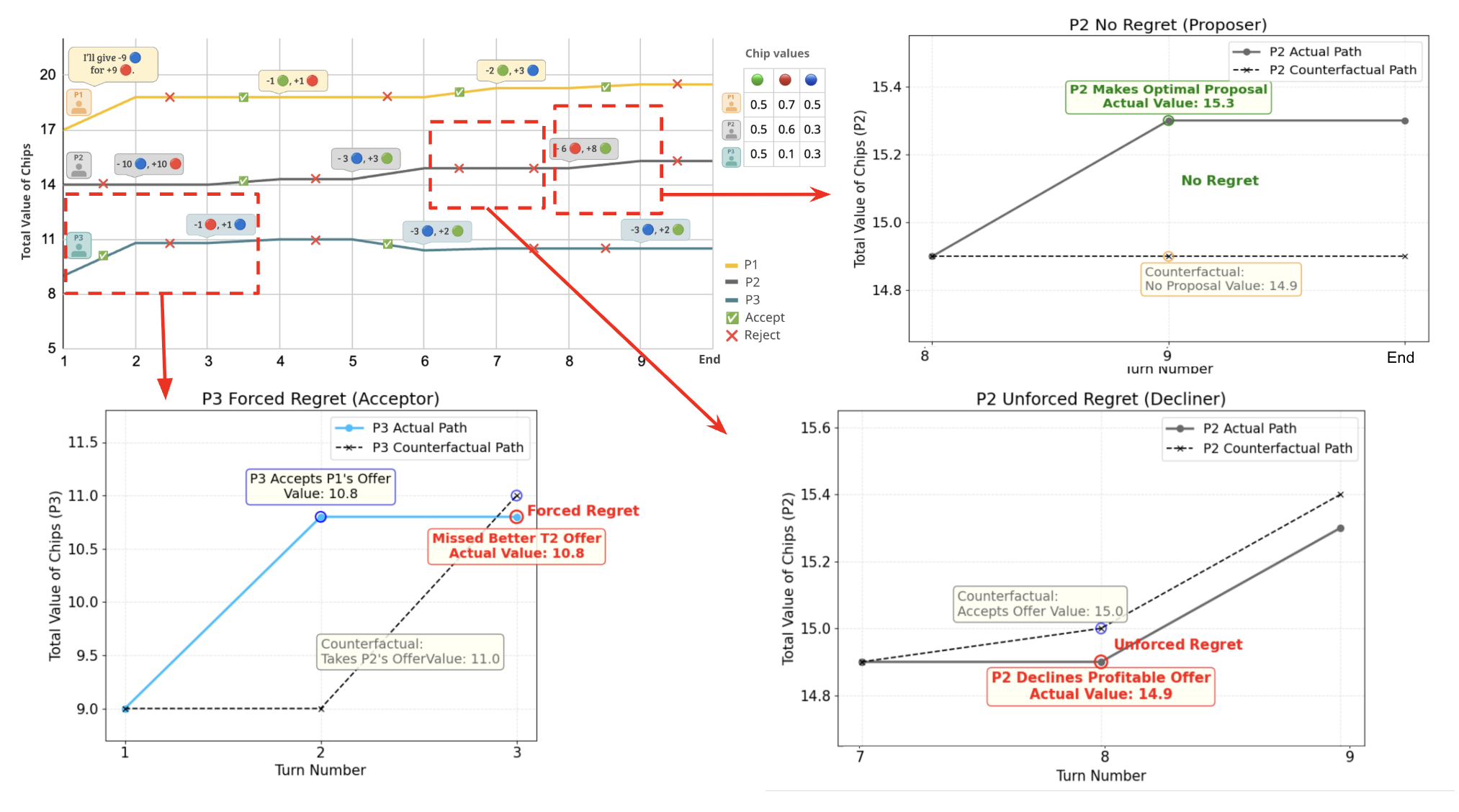}
\caption{Regret behavior visualization: The top-left panel shows an overview trajectory from Figure~\ref{fig:example-trajectory}. The top-right panel illustrates a ``No Regret'' scenario for P2 as a proposer. The bottom-left panel depicts ``Forced Regret'' for P3 as an acceptor. The bottom-right panel shows ``Unforced Regret'' for P2 as a decliner at Turn 8. Solid lines represent actual player paths, while dashed lines indicate counterfactual paths.}
\Description{Regret behavior visualization: The top-left panel shows an overview trajectory from Figure~\ref{fig:example-trajectory}. The top-right panel illustrates a ``No Regret'' scenario for P2 as a proposer. The bottom-left panel depicts ``Forced Regret'' for P3 as an acceptor. The bottom-right panel shows ``Unforced Regret'' for P2 as a decliner at Turn 8. Solid lines represent actual player paths, while dashed lines indicate counterfactual paths.}
\label{fig:regret_vis}
\end{figure*}

Here, we utilized \textit{counterfactual analysis} to visually dissect the three kinds of regret behavior previously discussed in Section~\ref{sec:regret}, as illustrated in Figure~\ref{fig:regret_vis}.
In these visualizations, the player's actual path of total chip value is depicted by a solid colored line, while the counterfactual path---representing the outcome had a different decision been made---is indicated by a gray dashed line.

\emph{Counterfactual definition:} The counterfactual path is constructed by simulating an alternative decision at a critical juncture (e.g., accepting a declined offer, declining an accepted offer, or making a different proposal) and projecting the subsequent value trajectory. This allows for a direct comparison between the actual outcome and what could have been.

\emph{No regret scenarios.}
This scenario is demonstrated by Player 2 (P2) in the role of a proposer, as shown in the upper-right panel of Figure~\ref{fig:regret_vis}, focusing on Turn 9. 
Player 2 makes a proposal that is accepted, leading to an actual chip value of $15.3$. The counterfactual path, representing a scenario where P2 might not have made this proposal or made a less optimal one, results in a lower value of 14.9. 
In this ``No Regret'' instance, the player's actual total value path is superior to or equal to the evaluated counterfactual path, indicating that the decision made was optimal given the available information and subsequent events. 

\emph{Forced regret scenarios.}
The lower-left panel of Figure~\ref{fig:regret_vis} illustrates ``Forced Regret'' experienced by Player 3 (P3) as an acceptor from Turn 1 to Turn 2. 
P3 initially accepts an offer from P1 at Turn 1, resulting in an actual value of 10.8. However, a more lucrative offer from P2 becomes available in Turn 2. 
Due to the commitment made in Turn 1 (e.g., changed chip inventory), P3 is unable to capitalize on this subsequent, better opportunity. The counterfactual path shows that had P3 declined P1's initial offer, they could have potentially accepted P2's offer, leading to a higher value of 11.0. 

\emph{Unforced regret scenarios.} The lower-right panel of Figure~\ref{fig:regret_vis} demonstrates Player 2 (P2) in the role of a decliner from Turn 8 to 9. 
P2 is presented with a profitable trade offer but chooses to decline it. 
P2's actual chip value remains at 14.9 after this decision. The counterfactual path, however, demonstrates that had P2 accepted this profitable offer, their chip value would have increased to 15.0. 
This highlights an avoidable error: a superior option was available and actionable, but the agent failed to select it, leading to a sub-optimal outcome. 
Unlike forced regret, no prior decision prevented P2 from taking the better option; the regret stems directly from the choice made at that specific juncture.

\section{Surplus Values and Significance}\label{app:results}

Surplus values are provided in Table~\ref{tab:summary_metric_table} and Table~\ref{tab:sig-surplus}.

\begin{table}[h]
\footnotesize
\centering
\begin{tabular}{lccc}
\toprule
Population & 2-chip & 3-chip & 4-chip \\
\midrule
Human           & 0.60 (0.06) & 0.59 (0.04) & 0.54 (0.03) \\
GPT-4o          & 0.69 (0.03) & 0.62 (0.03) & 0.54 (0.02) \\
GPT-4o refined  & 0.68 (0.04) & 0.64 (0.03) & 0.58 (0.02) \\
Gemini          & 0.42 (0.04) & 0.42 (0.03) & 0.37 (0.02) \\
Gemini refined  & 0.45 (0.05) & 0.43 (0.03) & 0.33 (0.02) \\
Bayesian agent  & 0.74 (0.04) & 0.80 (0.03) & 0.73 (0.02) \\
\bottomrule
\end{tabular}
\caption{This table provides the means and standard errors for the average ratio of observed total surplus to optimal surplus achieved across agent-game combinations. It corresponds to the numbers in Figure~\ref{fig:compare_agents_llm_2}.}
\label{tab:summary_metric_table}
\end{table}

\begin{table}[h!]
\centering
\scriptsize
\resizebox{\linewidth}{!}{%
\begin{tabular}{lcccccc}
\toprule
& Human & GPT-4o & GPT-4o refined & Gemini & Gemini refined & Bayesian agent \\
\midrule
\multicolumn{7}{l}{\textbf{2-chip}} \\
\midrule
Human & -- & 0.491 ↑ & 0.597 ↑ & 0.263 ↓ & 0.066 ↓ & 0.654 ↑ \\
GPT-4o & -- & -- & 0.302 ↓ & $p<0.001$ ↓ & $p<0.001$ ↓ & 0.141 ↑ \\
GPT-4o refined & -- & -- & -- & $p<0.001$ ↓ & $p<0.001$ ↓ & 0.414 ↑ \\
Gemini & -- & -- & -- & -- & 0.359 ↑ & $p<0.001$ ↑ \\
Gemini refined & -- & -- & -- & -- & -- & $p<0.001$ ↑ \\
Bayesian agent & -- & -- & -- & -- & -- & -- \\
\midrule
\multicolumn{7}{l}{\textbf{3-chip}} \\
\midrule
Human & -- & 0.525 ↑ & 0.021 ↑ & 0.001 ↓ & $p<0.001$ ↓ & $p<0.001$ ↑ \\
GPT-4o & -- & -- & 0.164 ↑ & $p<0.001$ ↓ & 0.004 ↓ & $p<0.001$ ↑ \\
GPT-4o refined & -- & -- & -- & $p<0.001$ ↓ & $p<0.001$ ↓ & 0.001 ↑ \\
Gemini & -- & -- & -- & -- & 0.799 ↑ & $p<0.001$ ↑ \\
Gemini refined & -- & -- & -- & -- & -- & $p<0.001$ ↑ \\
Bayesian agent & -- & -- & -- & -- & -- & -- \\
\midrule
\multicolumn{7}{l}{\textbf{4-chip}} \\
\midrule
Human & -- & 0.601 & 0.258 ↑ & $p<0.001$ ↓ & $p<0.001$ ↓ & $p<0.001$ ↑ \\
GPT-4o & -- & -- & 0.270 ↑ & $p<0.001$ ↓ & $p<0.001$ ↓ & $p<0.001$ ↑ \\
GPT-4o refined & -- & -- & -- & $p<0.001$ ↓ & $p<0.001$ ↓ & $p<0.001$ ↑ \\
Gemini & -- & -- & -- & -- & 0.003 ↓ & $p<0.001$ ↑ \\
Gemini refined & -- & -- & -- & -- & -- & $p<0.001$ ↑ \\
Bayesian agent & -- & -- & -- & -- & -- & -- \\
\bottomrule
\end{tabular}%
}
\caption{Two-sided t-test p-values comparing surplus values across populations (144 games). Arrows indicate whether the column's mean surplus was higher (↑) or lower (↓) than the row's.}
\label{tab:sig-surplus}
\end{table}

\newpage \clearpage

\section{Additional Data Analysis}
\subsection{Learning Effects}\label{app:learning}

\begin{figure}[ht!]
\centering
\includegraphics[width=\linewidth]{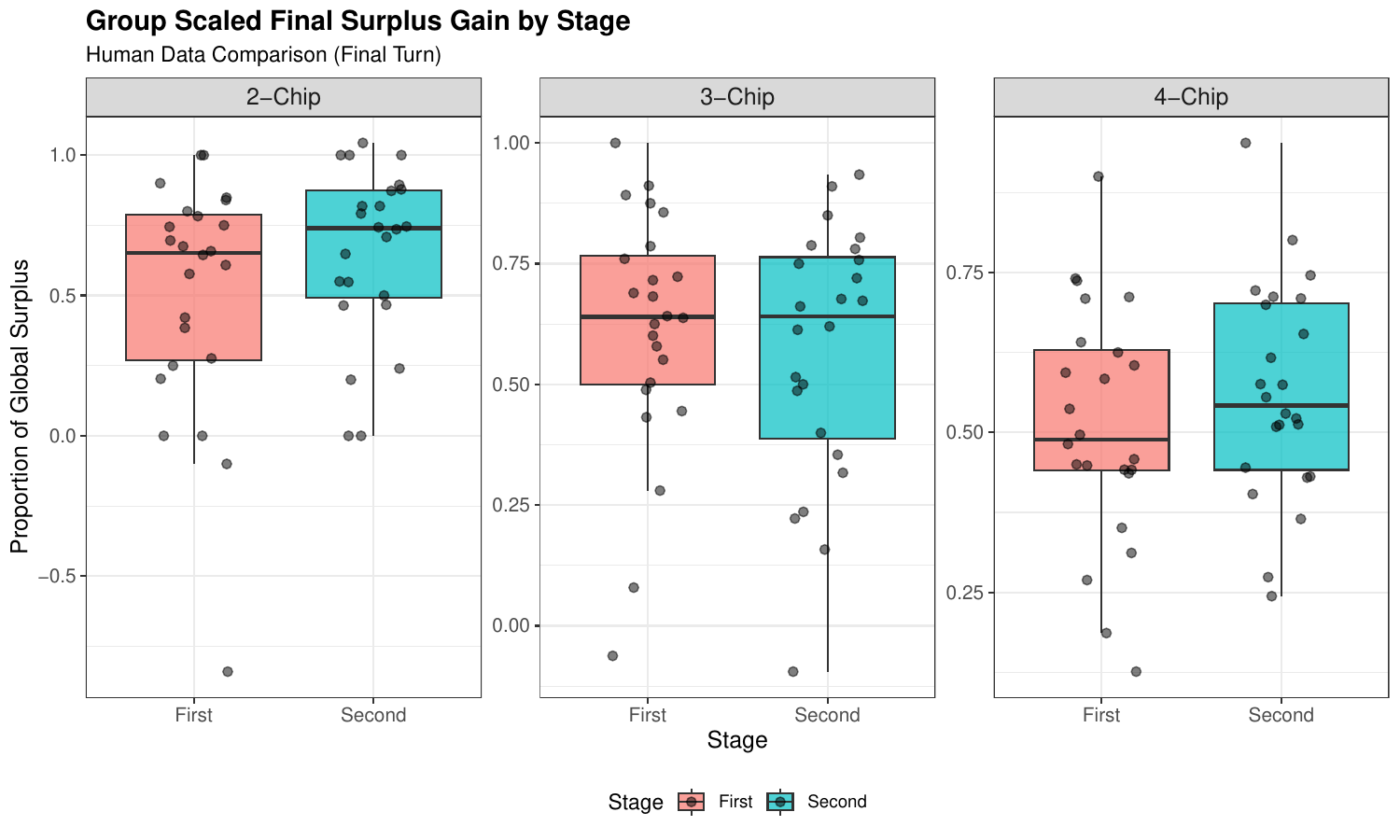}
\caption{Between stage (game) comparisons of human group surplus by game variant.}
\Description{Between stage (game) comparisons of human group surplus by game variant.}
\label{fig:learning-effect}
\end{figure}

In Figure \ref{fig:learning-effect}, we investigate learning effects for humans in the 2, 3, and 4-chip variants of the negotiation game by comparing the final surplus gain across the two stages of play. We see that while there is a small increase in performance in each game, it is not statistically significant.
We also cannot determine how the complexity of the game relates to learning effects, as the 3-chip game seems to have the smallest effect.

\subsection{Supplementary Trading Pattern Visualizations}\label{app:trade_space}

\begin{figure*}[ht!]
\centering
\includegraphics[width=0.8\linewidth]{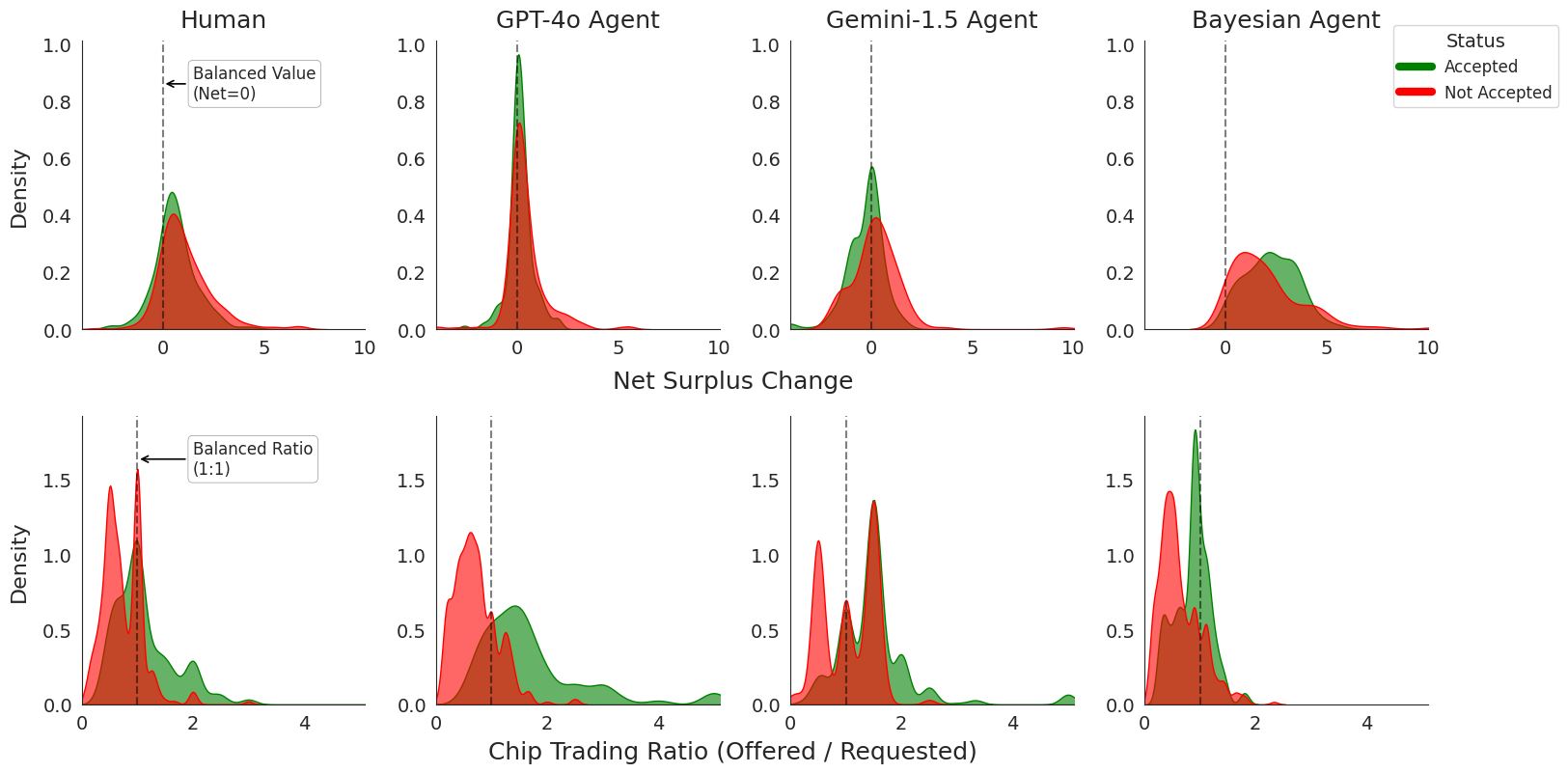}
\caption{This figure is identical to Figure~\ref{fig:trade-space} in the main text, but provides the trading patterns for the \emph{2-chip} game.}
\Description{This figure is identical to Figure~\ref{fig:trade-space} in the main text, but provides the trading patterns for the \emph{2-chip} game.}
\label{fig:trade_space2}
\end{figure*}

\begin{figure*}[ht!]
\centering
\includegraphics[width=0.8\linewidth]{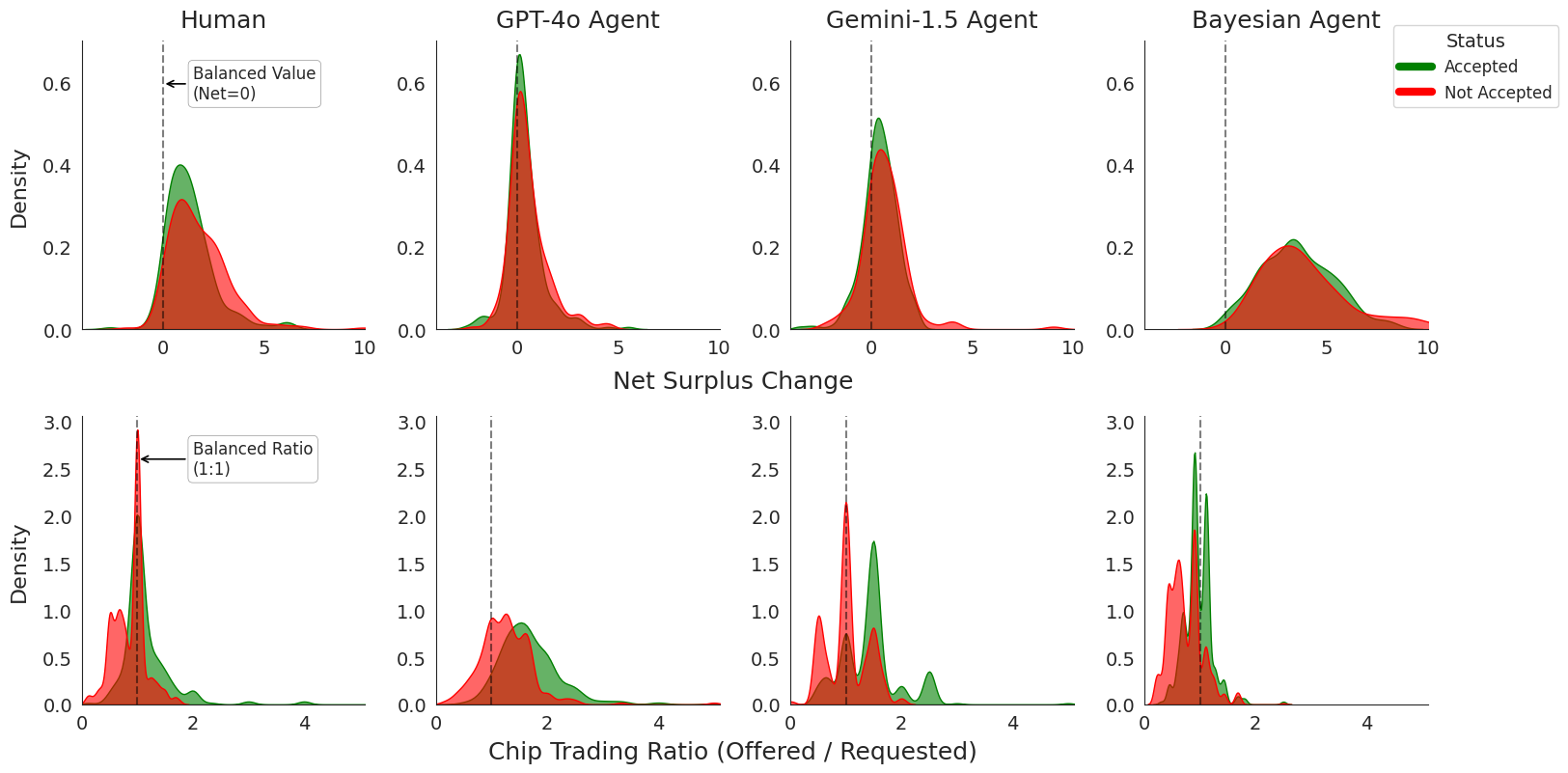}
\caption{This figure is identical to Figure~\ref{fig:trade-space} in the main text, but provides the trading patterns for the \emph{4-chip} game.}
\Description{This figure is identical to Figure~\ref{fig:trade-space} in the main text, but provides the trading patterns for the \emph{4-chip} game.}
\label{fig:trade_space4}
\end{figure*}

Summary statistics of trading patterns are provided in Table~\ref{tab:summary_trade_table}, and average final surplus information in Table~\ref{tab:summary_metric_table}. The 2-chip visualization is shown in Figure~\ref{fig:trade_space2} and 4-chip visualization in Figure~\ref{fig:trade_space4}. They exhibit similar behaviors as discussed in Section~\ref{sec:trade_patterns}.

\begin{table}[ht!]
\centering
\scriptsize
\resizebox{\linewidth}{!}{%
\begin{tabular}{llcccc}
\toprule
Game & Player & Acceptance & Surplus Mean (SD) & Ratio Mean (SD) & Median [Surplus, Ratio] \\
\midrule

\multirow{8}{*}{2-chip} 
& \multirow{2}{*}{Human} 
& Accepted   & 0.611 (1.031) & 1.147 (0.652) & [0.500, 1.000] \\
&              & Rejected   & 1.088 (1.313) & 0.735 (0.368) & [0.800, 0.667] \\
\cmidrule(lr){2-6}
& \multirow{2}{*}{GPT-4o} 
& Accepted   & 0.148 (0.617) & 1.711 (0.976) & [0.100, 1.500] \\
&              & Rejected   & 0.414 (1.031) & 0.715 (0.393) & [0.100, 0.667] \\
\cmidrule(lr){2-6}
& \multirow{2}{*}{Gemini-1.5 Pro} 
& Accepted   & -0.305 (0.913) & 1.508 (0.721) & [-0.100, 1.500] \\
&              & Rejected   & 0.230 (1.317) & 1.048 (0.475) & [0.100, 1.000] \\
\cmidrule(lr){2-6}
& \multirow{2}{*}{Bayesian} 
& Accepted   & 2.245 (1.252) & 0.867 (0.315) & [2.300, 0.900] \\
&              & Rejected   & 2.058 (1.904) & 0.636 (0.372) & [1.500, 0.526] \\

\midrule
\multirow{8}{*}{3-chip} 
& \multirow{2}{*}{Human} 
& Accepted   & 1.211 (1.317) & 1.039 (0.387) & [1.100, 1.000] \\
&              & Rejected   & 1.830 (1.803) & 0.840 (0.325) & [1.300, 1.000] \\
\cmidrule(lr){2-6}
& \multirow{2}{*}{GPT-4o} 
& Accepted   & 0.437 (0.838) & 1.765 (0.945) & [0.200, 1.600] \\
&              & Rejected   & 0.345 (0.963) & 0.918 (0.501) & [0.200, 0.800] \\
\cmidrule(lr){2-6}
& \multirow{2}{*}{Gemini-1.5 Pro} 
& Accepted   & 0.329 (0.899) & 1.394 (0.610) & [0.400, 1.500] \\
&              & Rejected   & 0.528 (1.819) & 1.075 (0.497) & [0.600, 1.000] \\
\cmidrule(lr){2-6}
& \multirow{2}{*}{Bayesian} 
& Accepted   & 3.173 (1.853) & 0.944 (0.274) & [3.200, 0.900] \\
&              & Rejected   & 3.385 (2.725) & 0.783 (0.355) & [2.800, 0.700] \\

\midrule
\multirow{8}{*}{4-chip} 
& \multirow{2}{*}{Human} 
& Accepted   & 1.331 (1.214) & 1.173 (0.562) & [1.200, 1.000] \\
&              & Rejected   & 1.757 (1.438) & 0.863 (0.288) & [1.500, 1.000] \\
\cmidrule(lr){2-6}
& \multirow{2}{*}{GPT-4o} 
& Accepted   & 0.369 (0.929) & 1.745 (0.831) & [0.200, 1.600] \\
&              & Rejected   & 0.534 (0.981) & 1.262 (0.518) & [0.300, 1.250] \\
\cmidrule(lr){2-6}
& \multirow{2}{*}{Gemini-1.5 Pro} 
& Accepted   & 0.403 (0.858) & 1.438 (0.540) & [0.400, 1.500] \\
&              & Rejected   & 0.650 (1.187) & 1.022 (0.377) & [0.600, 1.000] \\
\cmidrule(lr){2-6}
& \multirow{2}{*}{Bayesian} 
& Accepted   & 3.536 (1.793) & 0.965 (0.251) & [3.400, 0.900] \\
&              & Rejected   & 4.251 (2.824) & 0.738 (0.303) & [3.500, 0.700] \\

\bottomrule
\end{tabular}%
}
\caption{Summary statistics of the trading patterns in Figures~\ref{fig:trade-space}, \ref{fig:trade_space2}, and \ref{fig:trade_space4}.}
\label{tab:summary_trade_table}
\end{table}

\newpage \clearpage

\section{Aside on Smaller Model Performance}\label{app:smaller_models}

We tested smaller models from both the Gemini and GPT families---\texttt{Gemini-2.5-Flash} and \texttt{GPT-o4-mini}---and found that their overall performance falls short of both human participants and larger models such as GPT-4o, particularly in terms of surplus generation.
Even with prompt refinement, the average surplus remains substantially lower.

\begin{figure}[h!]
\centering
\includegraphics[width=\linewidth]{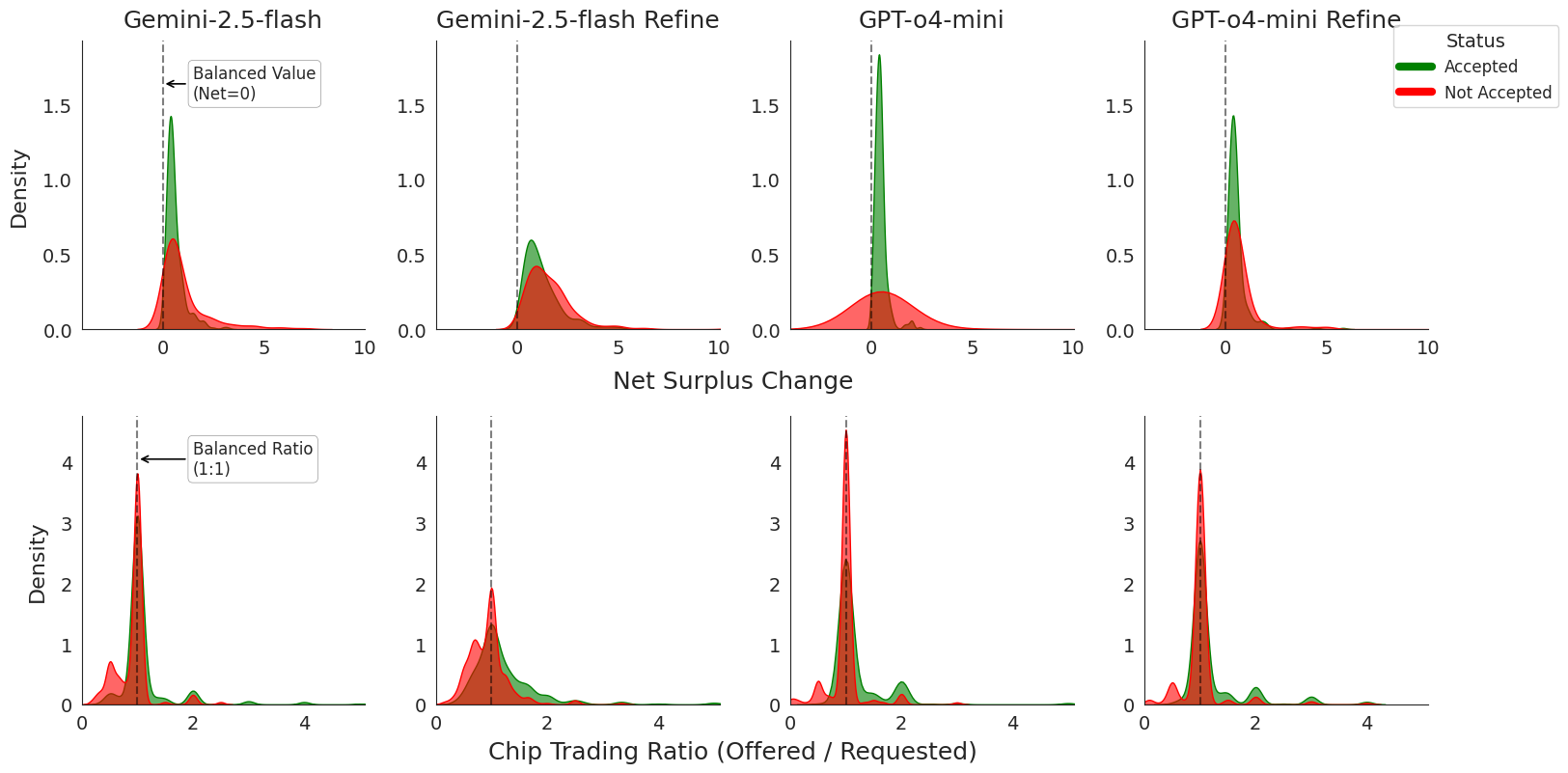}
\caption{This figure is identical to Figure~\ref{fig:trade-space} in the main text, but provides the trading patterns for the smaller models}
\Description{This figure is identical to Figure~\ref{fig:trade-space} in the main text, but provides the trading patterns for the smaller models}
\label{fig:small-model}
\end{figure}

Procedurally, these small models also behave quite differently from their larger counterparts. As shown in Figure~\ref{fig:small-model}, proposals by Gemini-2.5-Flash and GPT-o4-mini are heavily concentrated around a narrow region: nearly zero net surplus gain and a 1:1 trade ratio. Roughly 90\% of offers consist of exchanging one chip for exactly one of another type (e.g., 1 red for 1 green). Rarely do small LLMs propose trades that would result in a loss. However, they also rarely propose trades that would benefit them substantially (e.g., a net surplus of $> 2$). Prompt refinement slightly broadens the distribution but fails to shift the models away from this dominant 1:1 pattern.

This low-variance, risk-averse behavior is characteristic of the ``lazy generation'' effect reported in small language models \citep{lambert2023alignment}, and reflects the narrow, template-driven action spaces observed in recent bargaining benchmarks for sub-10B parameter agents \citep{xia-etal-2024-measuring}.

\section{LLM Simulation Prompts}\label{app:prompt}

\subsection{Rule Explanation}
\begin{lstlisting}[basicstyle=\ttfamily\tiny, breaklines=true]

How the game works:
The game consists of 3 rounds of trading. During each round, each player will have a turn to propose 1 trade. These turns are pre-determined in a random order, and the order stays the same in each round.

Trade proposals:
To propose a trade, a player must:
1. Request a certain quantity of chips of a single color from any other player to get.
2. Specify a certain quantity of chips of a different color to give in return.

Trade rules:
Players cannot offer more chips than they currently hold. For example, if you only have 5 red chips, you cannot offer 6 red chips.
Players cannot trade chips of the same color. For example, you cannot trade red chips for red chips.

Trade completion:
When an offer is presented, all other active participants get a chance to accept or decline. Note: Active participants are those not currently making the offer.

Participants make their decisions simultaneously and privately. The participant who receives the offer is not dependent on who accepts the trade first. Some possible outcomes:
If no one accepts, the trade does not happen, and the turn ends.
If multiple participants accept, one accepting participant is chosen at random to complete the trade with the offering participant. This means that participants cannot choose who they trade with.
If only one participant accepts, the trade will happen.

Key points to remember:
In each round, each player gets to propose one trade and respond to other players' trades
You can only propose trades between different colored chips, and cannot offer to give a chip amount that you do not have
When multiple players accept a trade, the trading partner is randomly selected
\end{lstlisting}

\subsection{Proposer's Prompt}

\begin{lstlisting} [basicstyle=\ttfamily\tiny, breaklines=true]

You are {{name}}.
Your valuations of the different types of chips are: {{preference_description}}.
You now have the following amounts of each chip: {{item}}.
The conversation history so far is {{history}}.

REMEMBER, to propose a trade, you must:
Request a certain quantity of chips of a single color from any other player to get
Specify a certain quantity of chips of a different color to give in return

REMEMBER you have the following amounts of each chip: {{item}}.
Your goal is to make as much money as possible. The trades, you choose to make to accomplish this, are up to you.
As a part of making money you must be rational - do not propose a trade in which you lose money. The value of a trade to you is the difference between the total value of chips you receive (quantity x your valuation) minus the total value of chips you give up (quantity x your valuation). Only propose trades that give you positive value.
In short, your trades should be both incentive compatible and individually rational.
Your response must use these EXACT tags below. The response should include nothing else besides the tags, your trade offer, and your reasoning. The text between tags should be concise.
```
<REASONING>
[Provide your concise reasoning in a few sentences, e.g. To gain more surplus, I want more xxx chips]
</REASONING>

<CHECK>
[check if you have sufficient chips to trade. If you have n green chips, you can at most give n green chips. If you don't want to trade, you can ask for a large amount of chips that no one can afford]
</CHECK>

<GET_COLOR> Color, e.g. red</GET_COLOR>
<GET_QUANTITY> quantity, e.g. n </GET_QUANTITY>
<GIVE_COLOR> Color, e.g. red</GIVE_COLOR>
<GIVE_QUANTITY> quantity, e.g. n </GIVE_QUANTITY>
```
\end{lstlisting}

\subsection{Receiver's Prompt}

\begin{lstlisting} [basicstyle=\ttfamily\tiny, breaklines=true]
You are {{name}}.
Your valuations of the different types of chips are: {{preference_description}}.
You now have the following amounts of each chip: {{item}}.
The conversation history so far is {{history}}.

You have an offer. {{proposer}} is offering to give {{give}} and get {{get}} in return.
If you make this trade, your total wealth will change by: {{delta_surplus}}

Now, you need to decide whether to accept or decline.
Your response must use these EXACT tags below. The response should include nothing else besides the tags, your choice to accept or decline, and your reasoning. The text between tags should be concise.
```
<REASONING>
[Provide your concise reasoning in a few sentences.]
</REASONING>

<CHOICE>Yes or No </CHOICE>
\end{lstlisting}

\subsection{Refined Proposing Prompt} \label{App:refine_prompt}

\subsubsection{Multiple proposal generation prompt}

\begin{lstlisting} [basicstyle=\ttfamily\tiny, breaklines=true]
You are {{name}}.
Your valuations of the different types of chips are: {{preference_description}}.
You now have the following amounts of each chip: {{item}}.
The conversation history so far is {{history}}.

REMEMBER, to propose a trade, you must:
Request a certain quantity of chips of a single color from any other player to get
Specify a certain quantity of chips of a different color to give in return

REMEMBER you have the following amounts of each chip: {{item}}.
Your goal is to make as much money as possible. The trades, you choose to make to accomplish this, are up to you.
As a part of making money you must be rational - do not propose a trade in which you lose money. The value of a trade to you is the difference between the total value of chips you receive (quantity $\times$ your valuation) minus the total value of chips you give up (quantity $\times$ your valuation). Only propose trades that give you positive value.
In short, your trades should be both incentive compatible and individually rational.
You can trade as many chips as you want in a single turn, assuming you have that many. Do not feel constrained to only trade a single chip at a time.
Propose 3 different good trade ideas, so a next step can decide on the best trade of the ones you propose here.
Your response must use these EXACT tags below. The response should include nothing else besides the tags, your trade offer, and your reasoning. The text between tags should be concise.
Repeat the below tags once for each of the trade ideas you propose.
```
<REASONING>
[Provide your concise reasoning in a few sentences, e.g. To gain more surplus, I want more xxx chips]
</REASONING>

<CHECK>
[check if you have sufficient chips to trade. If you have n green chips, you can at most give n green chips. If you don't want to trade, you can ask for a large amount of chips that no one can afford]
</CHECK>

<GET_COLOR> Color, e.g. red</GET_COLOR>
<GET_QUANTITY> quantity, e.g. n </GET_QUANTITY>
<GIVE_COLOR> Color, e.g. red</GIVE_COLOR>
<GIVE_QUANTITY> quantity, e.g. n </GIVE_QUANTITY>
```
\end{lstlisting}

\subsection{Choosing from Multiple Possible Proposals Prompt}

\begin{lstlisting} [basicstyle=\ttfamily\tiny, breaklines=true]
You are {{name}}.
Your valuations of the different types of chips are: {{preference_description}}.
You now have the following amounts of each chip: {{item}}.
The conversation history so far is {{history}}.

REMEMBER, to propose a trade, you must:
Request a certain quantity of chips of a single color from any other player to get
Specify a certain quantity of chips of a different color to give in return

REMEMBER you have the following amounts of each chip: {{item}}.
Your goal is to make as much money as possible. The trades, you choose to make to accomplish this, are up to you.
As a part of making money you must be rational - do not propose a trade in which you lose money. The value of a trade to you is the difference between the total value of chips you receive (quantity $\times$ your valuation) minus the total value of chips you give up (quantity $\times$ your valuation). Only propose trades that give you positive value.
In short, your trades should be both incentive compatible and individually rational.
You can trade as many chips as you want in a single turn, assuming you have that many. Do not feel constrained to only trade a single chip at a time.

Below are three trade ideas you have proposed. Please pick the best trade proposal from the ones below that you will propose to the group.
Do not change anything about the trade you have selected from the ideas below.
Your response must use these EXACT tags below. The response should include nothing else besides the tags and content of your selected trade offer including its reasoning.
```
<REASONING>
[Provide your concise reasoning in a few sentences, e.g. To gain more surplus, I want more xxx chips]
</REASONING>

<CHECK>
[check if you have sufficient chips to trade. If you have n green chips, you can at most give n green chips. If you don't want to trade, you can ask for a large amount of chips that no one can afford]
</CHECK>

<GET_COLOR> Color, e.g. red</GET_COLOR>
<GET_QUANTITY> quantity, e.g. n </GET_QUANTITY>
<GIVE_COLOR> Color, e.g. red</GIVE_COLOR>
<GIVE_QUANTITY> quantity, e.g. n </GIVE_QUANTITY>
```

Proposed trade ideas to choose from:
{{proposed}}
\end{lstlisting}

%%%%%%%%%%%%%%%%%%%%%%%%%%%%%%%%%%%%%%%%%%%%%%%%%%%%%%%%%%%%

\end{document}